\pdfoutput=1

\documentclass[11pt]{article}
\usepackage[final]{acl}
\usepackage{booktabs}
\usepackage{amsmath}
\usepackage{graphicx} 
\usepackage{subcaption} 
\usepackage{colortbl}
\usepackage{graphicx}
\usepackage{tabularx}
\usepackage{times}
\usepackage{listings}
\usepackage{latexsym}
\usepackage{float}
\usepackage{tcolorbox}
\usepackage{array, longtable, geometry}
\usepackage{framed}
\definecolor{boxbg}{RGB}{240,240,240}
\usepackage[T1]{fontenc}

\usepackage[utf8]{inputenc}

\usepackage{microtype}

\usepackage{inconsolata}

\usepackage{graphicx}
\definecolor{boxframe}{rgb}{0.5,0.5,0.5}

\newcommand{\ourmethod}{\textsc{WXImpactBench}}
\usepackage{xcolor}
\colorlet{green}{green!60!black}
\title{WXImpactBench: A Disruptive Weather Impact Understanding Benchmark for Evaluating Large Language Models}

\author{
 \textbf{Yongan Yu\textsuperscript{1}\thanks{Equal Contribution}},
 \textbf{Qingchen Hu\textsuperscript{1}\footnotemark[1]},
 \textbf{Xianda Du\textsuperscript{2}\footnotemark[1]},
  \textbf{Jiayin Wang\textsuperscript{3}},
  \textbf{Fengran Mo\textsuperscript{4}\thanks{Corresponding authors.}},
 \textbf{Renée Sieber\textsuperscript{1}\footnotemark[2]}
\\
\\
 \textsuperscript{1}McGill University,
 \textsuperscript{2}University of Waterloo,
   \textsuperscript{3}Tsinghua University,
  \textsuperscript{4}University of Montreal
\\
 {\tt yongan.yu@mail.mcgill.ca},
 {\tt fengran.mo@umontreal.ca},
 {\tt renee.sieber@mcgill.ca}
 }

\begin{document}
\maketitle
\begin{abstract}
Climate change adaptation requires the understanding of disruptive weather impacts on society, where large language models (LLMs) might be applicable. However, their effectiveness is under-explored due to the difficulty of high-quality corpus collection and the lack of available benchmarks. 
The climate-related events stored in regional newspapers record how communities adapted and recovered from disasters. However, the processing of the original corpus is non-trivial.
In this study, we first develop a disruptive weather impact dataset with a four-stage well-crafted construction pipeline. 
Then, we propose \ourmethod{}, the first benchmark for evaluating the capacity of LLMs on disruptive weather impacts.
The benchmark involves two evaluation tasks, multi-label classification and ranking-based question answering.
Extensive experiments on evaluating a set of LLMs provide first-hand analysis of the challenges in developing the understanding of disruptive weather impact and climate change adaptation systems.
The constructed dataset and the code for the evaluation framework are available to help society protect against disaster vulnerabilities.
\end{abstract}

\section{Introduction}
Climate change adaptation~\cite{karl1999climate}, referring to actions that help reduce societal vulnerability to climate change, demands a sophisticated understanding of the disruptive weather impacts on society (e.g., the perspective of economy and policy) \cite{carleton2016social}. Societal reactions to past disruptive weather events are stored in reliable historical sources \cite{cerveny2007extreme}.  
Among them, historical newspapers provide irreplaceable information, recording not just meteorological conditions \cite{gregory1981physical, brunet2011data}, but crucially, how societies adapted and recovered from disasters~\cite{norris2008community,handmer2012changes}. 
In addition, historical newspapers usually report regional disruptive weather impacts with local experiences, 
which is valuable to understanding long-term climate patterns and social effects \cite{ogilvie2010historical} but
are often absent in official meteorological records~\cite{vicky_challenge}.

Understanding complex patterns in disruptive weather events is important for society with forecasts, societal responses, and public policy~\cite{pielke2002weather}.
The challenge of identifying impacts and responses often lies in climate-related text processing, which contains period-specific narratives and domain-specific linguistic phenomena.
\begin{figure}[t!]
    \centering
    \includegraphics[width=\columnwidth]{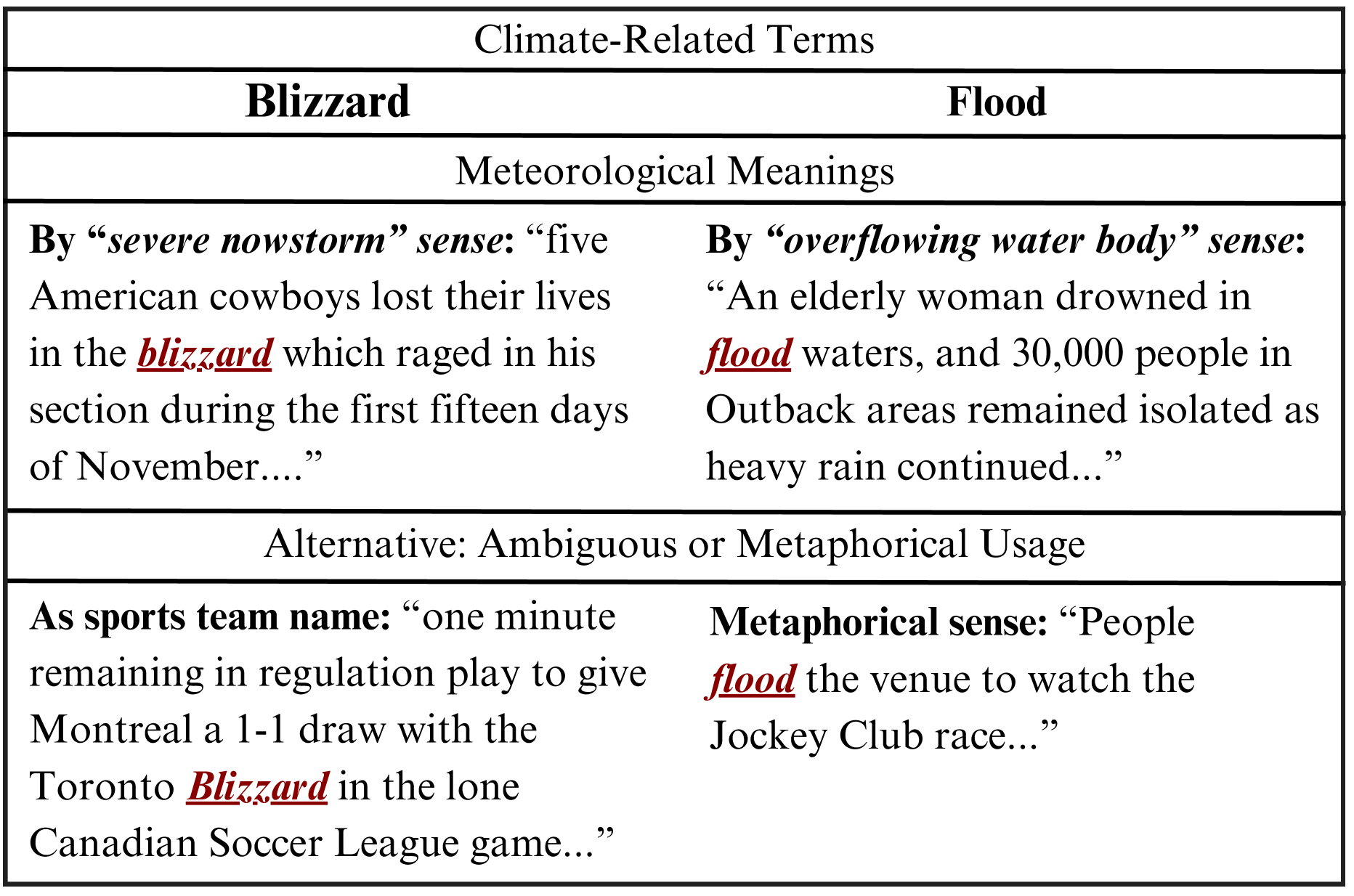}
    \caption{Climate-related polysemy examples in different narratives.}
\vspace{-2ex}
\label{weather_term}
\end{figure}
For example, disambiguation and taxonomy polysemy can occur in newspaper articles, where climate-related terms frequently appear in diverse linguistic contexts beyond their meteorological meanings.
Figure~\ref{weather_term} shows that the term ``blizzard'' can refer to a severe snowstorm or the name of the sports team (e.g., ``Toronto Blizzard''). Similarly, the term ``flood'' can describe an overflowing body of water or can be used metaphorically (e.g., ``flood the venue''). This polysemy occurs commonly in newspapers and thus requires the system to distinguish the literal weather-related meanings and alternate usages by improving the climate-related semantic understanding~\cite{nazeer2024evolution}.
Another challenge lies in extracting information from the original paper content. Although it is commonly achieved by optical character recognition (OCR)~\cite{thomas2024leveraging}, errors remain due to mixed content formats, and complex narrative structures~\cite{nazeer2024evolution}. These errors can negatively affect the extracted text for disruptive weather impact analysis, which renders the texts difficult to serve as a high-quality corpus.  

Existing studies on climate-related language processing focus on extracting climate patterns~\cite{alaparthi2020bidirectional}, wildfire resilience~\cite{xie2024wildfiregpt} and analyzing extreme weather events~\cite{mallick2024understanding}. 
Intuitively, LLMs ~\cite{tornberg2023use,mao2023large,yang2024harnessing,mo2024chiq} offer a powerful alternative for understanding disruptive weather impacts. 
However, their effectiveness is unexplored~\cite{boros-etal-2024-post,yuan2025omnigeo} due to the lack of a corresponding benchmark.
The resources used in previous studies cannot comprehensively evaluate the ability of LLMs for weather impacts.
This is because i) compared to informative reports in newspapers, previously used meteorological records do not contain long-term and detailed regional information~\cite{pevtsov2019historical}; 
and ii) the previous meteorological records are easily obtained and have been available for a long while. Thus they might be already included in the pre-training of LLMs and should not be included in benchmark build-up to avoid potential bias~\cite{Ferrara_2023}.
To develop a system that assesses the impact of disruptive weather on society, the first step is to establish a domain benchmark for the evaluation protocol.

In this study, we design a four-stage data construction pipeline that begins with a disruptive weather impact dataset in which we correct OCR errors in digitalized newspaper article extraction.
We extract a sample of articles from two time periods, which cover linguistic and social changes across different eras and increase linguistic complexity due to the different descriptions of weather events in different times~\cite {campbell2013historical}. Historical newspapers often employed more descriptive and elaborate narratives compared to modern reporting styles~\cite{bingham2010digitization}. These narratives frequently included outdated terminology, spelling variations, and evolving writing conventions \cite{campbell2013historical}.
The articles are selected by topic modeling, including six impact categories (infrastructural, political, financial, ecological, agricultural, and human health), which are informed by previous studies~\cite{imran_weather} and align with modern disaster impact assessment frameworks~\cite{silva2022building}.

With our constructed dataset, we develop a benchmark, \ourmethod{}, to investigate the capacity of LLMs to understand disruptive weather impacts with two tasks: i) multi-label classification and ii) ranking-based question-answering.
The multi-label classification task employs the previous six impact categories as labels for each article whose ground-truth is annotated by human labor.
The question and the candidate pools for the ranking-based question-answering task are constructed based on the context and annotation of the multi-label classification task. This can facilitate any future development of retrieval-augmented generation (RAG) systems in the climate-related domain \cite{zhao2024retrieval,mo2024survey,huang2024survey,de2025information}.
Extensive experiments on evaluating a set of off-the-shelf LLMs provide first-hand analysis of their capacity to understand disruptive weather impacts and reveal the challenges in developing climate change adaptation systems to help society protect against vulnerabilities from disasters.

Our contributions are summarized as follows:
(1) We construct a high-quality disruptive weather impact dataset from digitalized newspaper articles in the climate-related domain with a four-stage pipeline. 
(2) We propose \ourmethod{} with two typical tasks for evaluating the capacity of LLMs on disruptive weather impact understanding, which is the first benchmark to facilitate the development of such domain-specific systems. The constructed dataset and the evaluation framework code are available at our Github repository\footnote{\url{https://github.com/Michaelyya/WXImpactBench}}.
(3) We conduct extensive experiments on benchmarking a set of LLMs, providing first-hand analysis of challenges in disruptive weather impact understanding and climate change adaptation.

\begin{figure*}[t!]
    \centering
    \includegraphics[width=\textwidth]{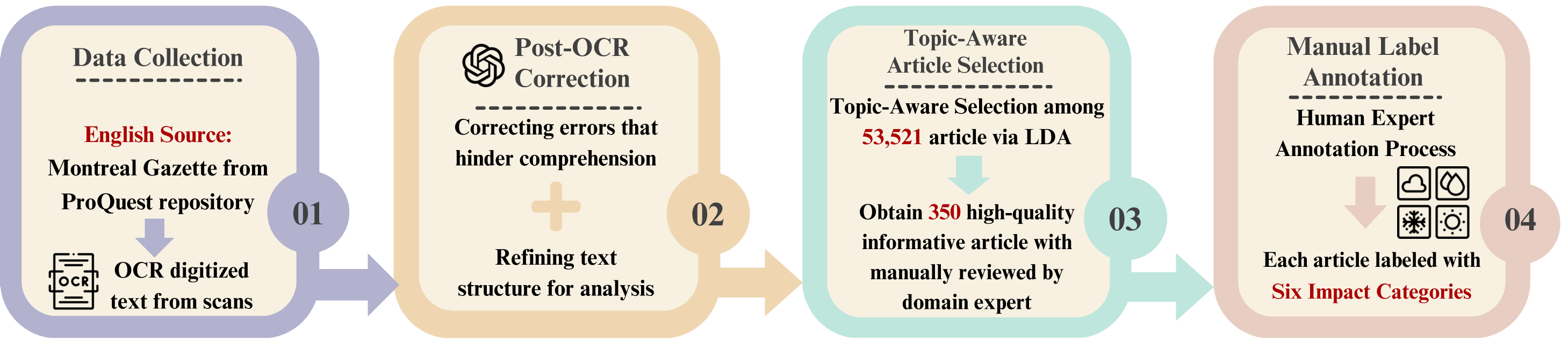}
    \caption{\textbf{Data Construction Pipeline} consists of four main stages: (1) Corpus collection from historical newspapers across two periods. (2) Post-OCR correction for high-quality extraction. (3) Article selection with defined categorization using LDA topic modeling and expert curation. (4) Annotation framework conducted by domain experts with a six-category impact classification scheme for understanding disruptive weather impacts.
    }
    \label{annotation_pipeline}
    \vspace{-2ex}
\end{figure*}

\section{Related Work}
\subsection{Climate Impact Analysis and Database}

Climate impact analysis \cite{thulke2024climategpt} aims 
to help society make correct decisions about climate-related challenges affecting communities, e.g., understanding the weather impacts on society.
Existing studies aim to validate the quality of historical weather data~\cite{renee2022trust} or extract climate patterns via name entities recognition tasks~\cite{mallick2024analyzing,xie2024wildfiregpt}.
Their used corpus is sourced from structured climate science materials, however, usually with daily loss~\cite{vicky_challenge}, due to the deterioration of storage media (paper, microfiche/microfilm, magnetic tape)~\cite{pevtsov2019historical}. 
Compared to structured climate-related scientific databases, historical newspapers can offer a better alternative due to their rich climate records \cite{vargas2021laclichev}, although they remain largely untapped \cite{krishnan2023climatenlp}. 
The scarcity of high-quality climate-related and nuanced textual data results in the lack of standard benchmarks, which limits understanding of weather impacts. We address these issues in this paper.

\subsection{Climate Text Processing and Benchmark}

Extracting and processing historical climate articles in newspapers is challenging due to their non-digital formats, such as scanned images or physical archives. OCR enables their conversion into machine-readable text \cite{baird2004difficult}, facilitating large-scale digitization, retrieval, and analysis. Like digital libraries \cite{singh2012survey}, OCR enhances accessibility, supporting research on climate trends and societal responses.
Although neural OCR correction models \cite{drobac2020optical} improve the quality of the extracted text, the degraded print quality, inconsistent terminology, and irregular column layouts \cite{binmakhashen2019document} cause potential errors, which negatively impact the text understanding and the usage for designing downstream tasks~\cite{bingham2010digitization,spathis2024first,wang2024user}. 

Thus, the lack of high-quality resources constrains the development of comprehensive benchmarks for weather impacts. 
\citet{li2024cllmate} introduce CLLMate, a multimodal benchmark that aligns meteorological data with textual event descriptions for weather event forecasting, though it focuses on prediction rather than historical impact understanding.
Developing a benchmark for understanding weather impacts is important, although the fragmented, incomplete, or dispersed disparate sources of weather events increase the difficulty of annotation~\cite{lamb2002climate,campbell2013historical}.
Although previous studies~\cite{Rasp_2020, rasp2024weatherbench2benchmarkgeneration} attempt to develop evaluation frameworks for physics-based weather forecasting models, they focus on data-driven weather modeling rather than weather impact understanding.
In this paper, we propose the first disruptive weather impact benchmark to fill in this blanks.

\section{WXImpactBench: Disruptive Weather Impact Benchmark}
Our \ourmethod{} benchmark aims to evaluate to what extent existing LLMs can understand disruptive weather impacts, which also shows the evolution of vulnerability and resilience strategies from society across various periods.
It involves two main stages: i) dataset construction; and ii) task definition and evaluation.

\subsection{Dataset Construction}
The construction of the dataset aims to obtain high-quality text samples.
The pipeline overview is presented in Figure~\ref{annotation_pipeline}, which consists of four stages: data collection, post-OCR correction, topic-aware article selection, and manual label annotation. \\

\noindent\textbf{Data Collection.} 
The data is obtained through collaboration with a proprietary archive institution covering two temporal periods.
The original data stored as digitalized text is obtained through OCR~\cite{cheriet2007character}, which contains substantial noise due to historical newspaper layouts, including uneven printing, varying font styles \cite{sulaiman2019degraded}, complex multicolumn structures \cite{binmakhashen2019document}, and overlapping text elements (e.g., advertisements) \cite{verhoef2015cultural}. 
Thus, post-OCR correction is necessary to ensure the corpus is high-quality ~\cite{chiron2017impact, traub2015impact}. Our choice of newspapers as the primary data source stems from their status as the most consistent and reliable documentation of social impacts from extreme weather events \citep{bingham2010digitization}. After the initial selection, we chose to examine a historical and a modern period, where the historical one was chosen as it had the greatest density of corresponding meteorological information, while the modern one is defined by the Intergovernmental Panel on Climate Change \citep{seneviratne2021weather}. \\

\noindent\textbf{Post-OCR Correction.} 
The goal of post-OCR correction is to correct errors that could significantly impact human comprehension or downstream task analysis, e.g., correct the inaccurate works split and remove unnecessary characters~\cite{o2013cleaning}.
For example, given the source image of a digital newspaper article (An example is provided in Appendix~\ref{OCR_example}), the text obtained by initial OCR and post-OCR correct is shown in Figure \ref{OCR-comparison}.
The post-OCR correction is achieved by deploying \textsc{GPT-4o} with customized prompts~\cite{gpt4ocr}. The correction quality is evaluated using BLEU and ROUGE metrics, achieving high consistency scores compared to human annotations, see Appendix~\ref{OCR_evaluation} for detailed evaluation results. Our specific prompts for correction are provided in Appendix~\ref{OCR_instruct}. \\

\noindent\textbf{Topic-Aware Article Selection.} After the post-OCR correction, we obtain 53,521 weather event-related articles. We aim to obtain informative samples across historical and modern periods based on weather categories.
This is achieved by conducting topic modeling on the article collection, where we categorize them via Latent Dirichlet Allocation (LDA)~\cite{blei2003latent} to obtain the topic words - representing the primary weather event categories. 
The details of the categories are provided in Appendix~\ref{Annotation_guideline}.
Selected articles with informative weather content are manually reviewed by three domain experts, which result in 350 high-quality samples.
This process ensures the selected articles are topic-aware, which is highly related to specific disruptive weather events. \\

\begin{figure}[t!]
    \small

    \medskip
    \noindent\textbf{\small Initial OCR-Digitized Text} \hrulefill
    
    \small
    \noindent  ' ', 't 1', ' v MK ' A 0. P. R. 2NGIlNE IN THE DIT0H. ' I ', "Th'- snow storm, which was a LL day on Mondy hover' ing overhead, began to set in at dusk, and it gradually increased in Severity, COntInuIng until abt six o-''clock yesterday morning. The storm was the w0rst for many wrintrs, and to find a preedent for it severity IT ''1 is said we h' ve to go back to the eventful year ...

    \medskip
    \noindent\textbf{\small After Post-OCR Correction} \hrulefill
    
    \small
    \noindent A C. P. R. ENGINE IN THE DITCH. The snow storm, which was all day on Monday hovering overhead, began to set in about dusk, and it gradually increased in severity, continuing until about six o'clock yesterday morning. The storm was the worst for many winters, and to find a precedent for it in severity it is said we have to go back to the eventful year ...
    \caption{Example of the text obtained from initial OCR and after our post-OCR correction.}
    \label{OCR-comparison}
\vspace{-2ex}
\end{figure}

\noindent\textbf{Manual Label Annotation.}
Having selected article samples, the next step is to assign the annotation for each of them based on the label space. 
Six vulnerability-related disruptive weather impacts are defined as the labeling categories, including Infrastructural, Political, Financial, Ecological, Agricultural, and Human Health.
Annotation is conducted by three domain annotators following our guidelines (provided in Appendix~\ref{Annotation_Instruction}). 
According to the guidelines, the annotators should assign binary labels to indicate the presence or absence of direct descriptions of specific impacts within each article. Unlike previous study~\cite{imran_weather}, however, each sample might correspond to more than one impact.

\subsection{Task Definition and Evaluation}
\begin{figure*}[t!]
    \centering
    \includegraphics[width=\textwidth]{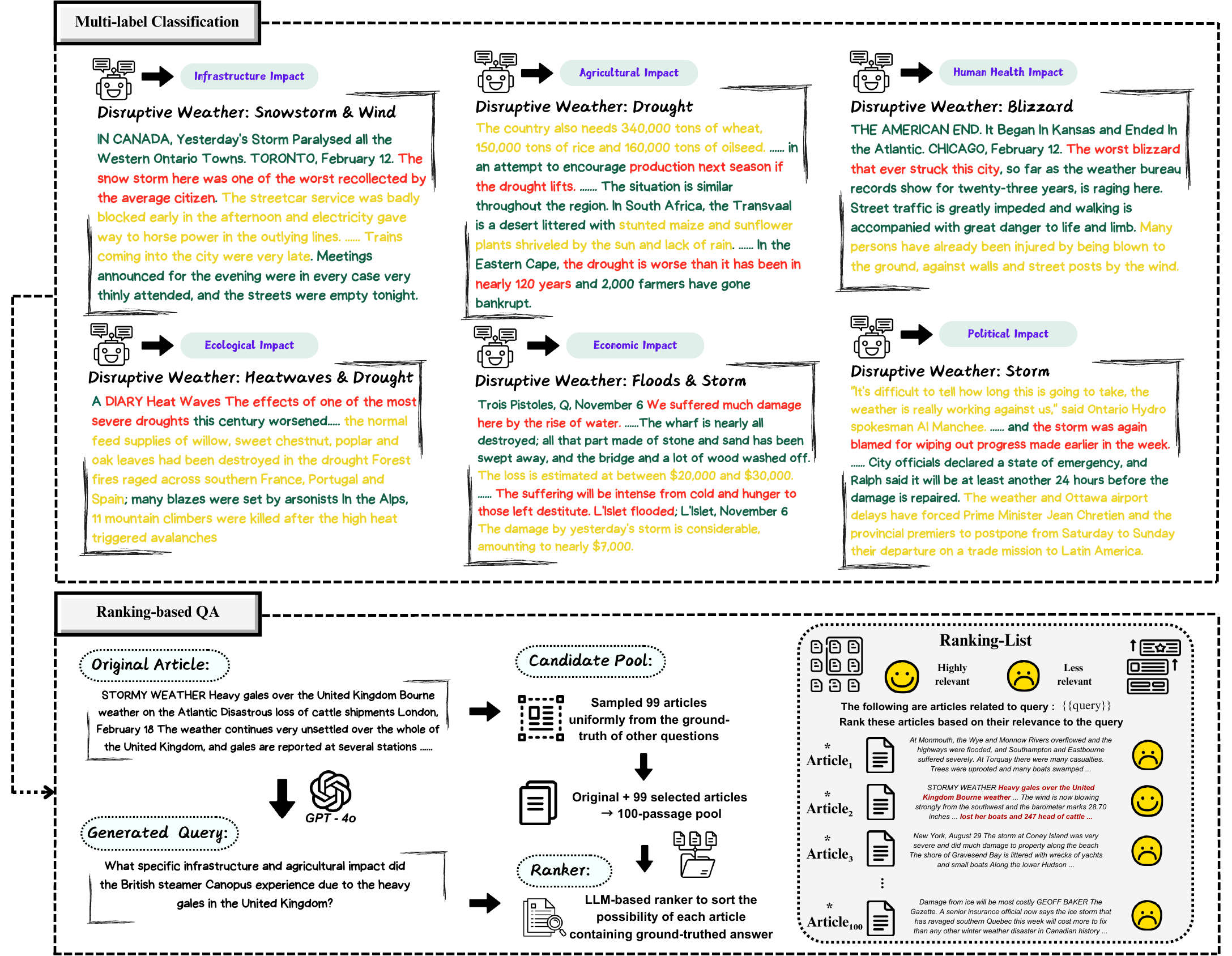}
    \caption{The overview of the benchmark framework with two tasks. Six disruptive weather impacts are used as labeling space in the classification task, where the \textcolor{red!85!black}{Red} texts represent \textbf{disruptive weather events} \textit{(e.g., snowstorm, drought, and blizzard)}, \textcolor{yellow!10!orange!}{Yellow} texts highlight \textbf{impact descriptions} \textit{(e.g., damage assessments, resource needs)}, and \textcolor{green!40!black}{Green} texts refer to \textbf{narrative descriptions} \textit{(e.g., geographical locations)}.
}
\vspace{-2ex}
    \label{impacts-annotation}
\end{figure*}

After finalizing the data construction, we design the evaluation framework for our benchmark \ourmethod{}. 
The overview is shown in Figure~\ref{impacts-annotation}, which contains two tasks, multi-label classification and ranking-based question-answering, to evaluate the capacity of LLMs to understand disruptive weather impacts.

\subsubsection{Multi-Label Text Classification}
With the annotated weather impact category for each selected article, the intuitive evaluation task is multi-label classification, which aims to test the ability of LLMs to distinguish the disruptive weather impact for each given article. 
The previous classification tasks in disaster-related natural language processing \cite{Purohit_Castillo_Diaz_Sheth_Meier_2013,imran-etal-2016-twitter} usually focus on modern crisis communications with structured text. 
Different from them, our constructed samples require the models to understand the linguistic shifts between historical and modern texts and address inconsistent styles of narratives across various periods. 
Specifically, given an article sample $x_t$ corresponding to six ground-truth impacts $\mathcal{Y}_t = \{y_t^i\}_{i=1}^6$ with binary labels ${y}_t^i \in \{0, 1\}$,
the evaluated model $\mathcal{M}$ is required to maximize the probability of the predicated impact $\hat{\mathcal{Y}}_t = \{\hat{y}_t^i\}_{i=1}^6$ towards ground-truth.
The objective function $\mathcal{L}$ for the given sample $x_t$ of multi-label classification task is formulated as 
$$
\mathcal{L}(\hat{\mathcal{Y}}_t, \mathcal{Y}_t) = -\sum_{i=1}^6 y_i\log \hat{y}_i, \quad \hat{y}_i = \mathcal{M}(x_t)
$$
\raggedbottom
\subsubsection{Ranking-based Question Answering}
Question-answering (QA) requires the LLMs to reply to the given question based on their parametric knowledge.
We formulate the ranking-based QA task by prompting the models to identify the likelihood of each article containing the correct answer from a candidate pool. 
This setting could also facilitate RAG systems development in the domain~\cite{mao2024rag,mo2024leverage,mo2025uniconv}, where we left the answer span extraction/generation for future studies.

To construct an evaluation protocol, the first step is to obtain suitable question pairs with each annotated samples in the multi-label classification task, since the question set is unavailable.
Thus, we generate pseudo questions $q_t$ for each article $(x_t, \mathcal{Y}_t)$ based on its annotated category via a generative LLM $\mathcal{G}$ which is formulated as
$q_t = \mathcal{G}(x_t, \mathcal{Y}_t)$.
The annotated categories $\mathcal{Y}_t$, which are the societal impacts brought by the disruptive weather event, will become part of the prompt to ensure the generated question targets one of the specific impact categories (see Figure~\ref{impacts-annotation}). 

As a result, we have QA pair $(q_t, x_t)$ for each sample. The next step is to construct the candidate pool for ranking. The size of the pool $\mathcal{X}_t$ for each question $q_t$ is $100$, which contains the ground-truth $x_t$ and other $99$ counterexamples $\mathcal{X}_t^{-}$ that are randomly sampled from the ground-truth of other questions. 
With the constructed QA pairs and corresponding candidate pools, the evaluated model $\mathcal{M}$ is required to rank the ground-truth based on the relevant scores produced by a matching function $\mathcal{S}$. The task objective can be formulated as 
$$
\text{RankingList} \ \phi(q_t) \leftarrow \mathop{\arg\max}_{\mathcal{S}(q_t, x_t)} \ \mathcal{M}(q_t, \mathcal{X}_t, \mathcal{S})
$$
\raggedbottom
where $\{x_t, \mathcal{X}_t^{-}\} \subseteq \mathcal{X}_t$ and the output ranking list $\phi(q_t)$ is evaluated by ranking metrics.

\section{Experimental Setup}

\subsection{Evaluation Metrics and Settings}
\label{sec:Evaluation}
For multi-label classification task, we use F1-score, accuracy, and row-wise accuracy as evaluation metrics. 
The evaluation via F1-score and accuracy are averaged across the six impact categories, historical and modern articles, and the effect of different context lengths.
Compared to the common F1-score and accuracy, the row-wise accuracy is a strict metric that requires more accurate output as the model should correctly classify all six impact labels for a given article, defined as  
$$
\text{Row-wise Acc.} = \frac{1}{N} \sum_{i=1}^{N} \prod_{j=1}^{6} \mathcal{I} \left( \hat{y}_i^j = y_i^j \right)
$$
\raggedbottom
where \( N \) is the number of samples, \( \hat{y}_i^j \) denotes the predicted label for the \( j \)-th category in the \( i \)-th sample, \( y_i^j \) is the corresponding ground-truth label, and \( \mathcal{I}(\cdot) \) is the indicator function.
The evaluation goal is to investigate the models' ability to distinguish various disruptive weather impacts under different settings, e.g., different periods.
For the long-context impact evaluation, we create an alternate version (mixed context), whose sample length is split into segments with approximately 250 tokens from the original one (long-context version) following \cite{levy-etal-2024-task}.
Note that annotations for these smaller chunks are performed independently by the same domain experts rather than automatically inherited from the original articles. 
This independent annotation process naturally results in some chunks containing no weather impact labels, which serve as valuable negative examples in our evaluation. These negative samples allow us to assess models' understanding to correctly identify passages without weather impacts, particularly reflected in our row-wise accuracy metric.
Eventually, we contain 350 and 1,386 samples for the original and mixed context version datasets, respectively. 
The detailed statistics of the datasets are provided in Appendix \ref{token-length}.

For the ranking-based QA task, we deploy the standard metric that emphasizes the accuracy of top positions for evaluation, including Hit@1, nDCG@5, Recall@5, and MRR.
\begin{table*}[h!]
\centering
\resizebox{\textwidth}{!}{
\begin{tabular}{lccccccc}
\hline
~ & \textbf{Infrastructure} & \textbf{Political} & \textbf{Financial} & \textbf{Ecological} & \textbf{Agricultural} & \textbf{Human Health} & \textbf{Average} \\ 

\hline
Positive Cases & 168 (326) & 61 (101) & 98 (134) & 54 (71) & 80 (100) & 117 (178) & -\\

\hline
\multicolumn{8}{c}{\textbf{Zero-Shot Performance}} \\
\hline
\textsc{GPT-4o} & \underline{80.94} \textcolor{green}{$\uparrow$ 0.39} & \textbf{58.46} \textcolor{green}{$\uparrow$ 3.70} & \textbf{65.82} \textcolor{red}{$\downarrow$ 0.23} & \textbf{46.81} \textcolor{green}{$\uparrow$ 2.51} & \textbf{70.33} \textcolor{green}{$\uparrow$ 1.92} & \textbf{73.23} \textcolor{green}{$\uparrow$ 1.12} & \textbf{65.93} \textcolor{green}{$\uparrow$ 1.07} \\

\textsc{GPT-4} & 74.87 \textcolor{green}{$\uparrow$ 2.13} & \underline{49.38} \textcolor{green}{$\uparrow$ 2.47} & 55.70 \textcolor{green}{$\uparrow$ 1.18} & 37.84 \textcolor{green}{$\uparrow$ 4.71} & 60.00 \textcolor{green}{$\uparrow$ 3.83} & 62.96 \textcolor{green}{$\uparrow$ 3.71} & 
56.79 \textcolor{green}{$\uparrow$ 2.12}\\ 

\textsc{GPT-3.5-turbo} & 77.59 \textcolor{green}{$\uparrow$ 3.73} & 41.60 \textcolor{green}{$\uparrow$ 7.73} & 42.39 \textcolor{green}{$\uparrow$ 7.16} & 36.52 \textcolor{red}{$\downarrow$ 0.01} & 55.63 \textcolor{green}{$\uparrow$ 4.49} & 47.29 \textcolor{green}{$\uparrow$ 10.91} &
50.17 \textcolor{green}{$\uparrow$ 5.12} \\ 

\hline
\textsc{DeepSeek-V3-671B} & \textbf{81.87} \textcolor{green}{$\uparrow$ 0.80} & 44.44 \textcolor{green}{$\uparrow$ 12.40} & \underline{60.91} \textcolor{green}{$\uparrow$ 3.91} & 36.00 \textcolor{red}{$\downarrow$ 0.60} & 61.74 \textcolor{green}{$\uparrow$ 4.34} & 65.20 \textcolor{green}{$\uparrow$ 0.20} &
58.03 \textcolor{green}{$\uparrow$ 3.07}
\\
\textsc{Mistral-24B-IT} & 79.12 \textcolor{red}{$\downarrow$  0.17}&	47.18 \textcolor{green}{$\uparrow$ 6.91}	&59.64 \textcolor{green}{$\uparrow$  2.04}	& \underline{44.90} \textcolor{green}{$\uparrow$ 10.75}	& \underline{67.74} \textcolor{green}{$\uparrow$ 1.07}	& \underline{66.88} \textcolor{green}{$\uparrow$  1.30}& \underline{60.91} \textcolor{green}{$\uparrow$ 1.57 }
\\ 
\textsc{Mixtral-8x7B-IT} & 72.31 \textcolor{green}{$\uparrow$ 3.01} & 39.29 \textcolor{green}{$\uparrow$ 6.86} & 57.02 \textcolor{green}{$\uparrow$ 0.33} & 36.59 \textcolor{green}{$\uparrow$ 5.23} & 44.44 \textcolor{green}{$\uparrow$ 14.23} & 50.00 \textcolor{green}{$\uparrow$ 9.69} &
49.94 \textcolor{green}{$\uparrow$ 4.84}
\\ 
\textsc{Mistral-7B-IT} & 74.27 \textcolor{green}{$\uparrow$ 4.27} & 36.63 \textcolor{green}{$\uparrow$ 6.63} & 45.56 \textcolor{green}{$\uparrow$ 0.56} & 39.19 \textcolor{green}{$\uparrow$ 5.19} & 55.30 \textcolor{green}{$\uparrow$ 5.30} & 61.61 \textcolor{green}{$\uparrow$ 11.61}&
52.76 \textcolor{green}{$\uparrow$ 1.98 }
\\ 
\textsc{Qwen2.5-14B-IT} & 76.21\textcolor{red}{$\downarrow$ 1.45} & 41.45 \textcolor{green}{$\uparrow$ 0.34} & 45.07 \textcolor{green}{$\uparrow$ 5.32} & 41.62 \textcolor{red}{$\downarrow$ 2.15} & 52.99 \textcolor{green}{$\uparrow$ 5.01} & 63.08 \textcolor{red}{$\downarrow$ 1.64} & 53.74 \textcolor{red}{$\downarrow$ 1.36}
\\ 
\textsc{Qwen2.5-7B-IT} & 70.52 \textcolor{green}{$\uparrow$ 4.75} & 34.29 \textcolor{green}{$\uparrow$ 2.82} & 43.43 \textcolor{green}{$\uparrow$ 0.83} & 41.06 \textcolor{red}{$\downarrow$ 6.51} & 40.26 \textcolor{green}{$\uparrow$ 3.96} & 38.19 \textcolor{green}{$\uparrow$ 7.34} &
 44.63 \textcolor{green}{$\uparrow$ 1.88 }
\\ 
\textsc{Gemma-2-9b-IT} & 77.42 \textcolor{green}{$\uparrow$ 1.52}	& 43.33 \textcolor{green}{$\uparrow$  2.59}& 	54.60 \textcolor{red}{$\downarrow$0.33}	&  42.16 \textcolor{green}{$\uparrow$  1.73}& 	55.60 \textcolor{green}{$\uparrow$  4.10}	& 61.82\textcolor{green}{$\uparrow$ 0.87} & 55.82\textcolor{green}{$\uparrow$ 1.31}
\\
\textsc{LLaMa-3.1-8B-IT} & 70.13 \textcolor{green}{$\uparrow$ 8.82} & 40.47 \textcolor{green}{$\uparrow$ 1.51} & 55.29 \textcolor{red}{$\downarrow$ 2.08} & 33.90 \textcolor{green}{$\uparrow$ 4.99} & 55.49 \textcolor{green}{$\uparrow$ 5.05} & 50.68 \textcolor{green}{$\uparrow$ 11.72} &
50.66 \textcolor{green}{$\uparrow$ 4.91 }
\\ 
\rowcolor{gray!20}
Average & 75.93 \textcolor{green}{$\uparrow$ 2.93} & 43.32 \textcolor{green}{$\uparrow$ 5.36} & 53.22 \textcolor{green}{$\uparrow$ 1.34} & 43.33 \textcolor{green}{$\uparrow$ 2.80} & 56.32 \textcolor{green}{$\uparrow$ 4.83} & 58.36 \textcolor{green}{$\uparrow$ 5.85} & 54.48 \textcolor{green}{$\uparrow$ 2.38} \\
\hline
\multicolumn{8}{c}{\textbf{In-Context Learning (One-shot) Performance}} \\
\hline
\textsc{GPT-4o} & \underline{81.25} \textcolor{green}{$\uparrow$ 0.29}& \textbf{59.54} \textcolor{green}{$\uparrow$ 2.71} & \textbf{63.64} \textcolor{green}{$\uparrow$ 1.18} & \textbf{50.00} \textcolor{red}{$\downarrow$ 0.02} & \textbf{71.43} \textcolor{green}{$\uparrow$ 1.39} & \textbf{72.94} \textcolor{green}{$\uparrow$ 2.80} & \textbf{66.93} \textcolor{green}{$\uparrow$ 1.02} \\ 
\textsc{GPT-4} & 72.63 \textcolor{green}{$\uparrow$ 3.54} & 40.00 \textcolor{green}{$\uparrow$ 5.92} & 55.15 \textcolor{green}{$\uparrow$ 0.55} & 32.38 \textcolor{green}{$\uparrow$ 7.30} & 61.29 \textcolor{green}{$\uparrow$ 2.92} & 60.22 \textcolor{green}{$\uparrow$ 5.06} & 
54.95\textcolor{green}{$\uparrow$ 3.25}\\
\textsc{GPT-3.5-turbo} & 76.88 \textcolor{green}{$\uparrow$ 2.29} & 38.93 \textcolor{green}{$\uparrow$ 9.72} & 48.50 \textcolor{green}{$\uparrow$ 0.95} & 40.00 \textcolor{green}{$\uparrow$ 0.02} & 57.30 \textcolor{red}{$\downarrow$ 0.02} & 60.98 \textcolor{red}{$\downarrow$ 0.02} & 
54.08\textcolor{green}{$\uparrow$ 2.26}\\  
\hline
\textsc{DeepSeek-V3-671B} & \textbf{81.62} \textcolor{green}{$\uparrow$ 1.32} & \underline{49.48} \textcolor{green}{$\uparrow$ 7.40} & \underline{63.37} \textcolor{green}{$\uparrow$ 2.10} & 43.55 \textcolor{green}{$\uparrow$ 1.21} & \underline{62.82} \textcolor{green}{$\uparrow$ 5.04} & \underline{67.78} \textcolor{green}{$\uparrow$ 3.07} & 
\underline{62.90} \textcolor{green}{$\uparrow$ 2.36} \\ 
\textsc{Mistral-24B-IT}& 78.38 \textcolor{green}{$\uparrow$ 0.58} & 43.48 \textcolor{green}{$\uparrow$ 12.36} & 56.99 \textcolor{green}{$\uparrow$ 1.05} & 35.09 \textcolor{green}{$\uparrow$ 6.39} & 61.45 \textcolor{green}{$\uparrow$ 4.50} & 65.28 \textcolor{red}{$\downarrow$ 0.08} & 57.41\textcolor{green}{$\uparrow$ 2.06} \\ 
\textsc{Mixtral-8x7B-IT} & 68.31 \textcolor{green}{$\uparrow$ 4.47} & 12.50 \textcolor{green}{$\uparrow$ 24.14} & 42.00 \textcolor{green}{$\uparrow$ 8.43} & 26.45 \textcolor{green}{$\uparrow$ 7.74} & 36.80 \textcolor{green}{$\uparrow$ 10.26} & 46.46 \textcolor{green}{$\uparrow$ 14.82} & 40.53\textcolor{green}{$\uparrow$ 8.63}\\ 
\textsc{Mistral-7B-IT}& 73.31 \textcolor{green}{$\uparrow$ 3.31} & 20.74 \textcolor{green}{$\uparrow$ 6.74} & 45.33 \textcolor{green}{$\uparrow$ 5.33} & 31.94 \textcolor{green}{$\uparrow$ 1.94} & 52.77 \textcolor{green}{$\uparrow$ 2.77} & 54.87 \textcolor{green}{$\uparrow$ 4.87} & 	47.43\textcolor{green}{$\uparrow$ 1.57}\\ 
\textsc{Qwen2.5-14B-IT}& 78.10 \textcolor{green}{$\uparrow$ 0.05} & 43.36 \textcolor{green}{$\uparrow$ 1.21} & 48.42 \textcolor{green}{$\uparrow$ 6.13} & \underline{43.65} \textcolor{green}{$\uparrow$ 0.30} & 62.18 \textcolor{green}{$\uparrow$ 4.49} & 63.60 \textcolor{green}{$\uparrow$ 3.78}& 54.95 \textcolor{green}{$\uparrow$ 1.71}  \\
\textsc{Qwen2.5-7B-IT}& 71.04 \textcolor{green}{$\uparrow$ 2.70} & 31.46 \textcolor{red}{$\downarrow$ 4.19} & 48.80 \textcolor{green}{$\uparrow$ 0.69} & 37.68 \textcolor{red}{$\downarrow$ 0.09} & 47.54 \textcolor{red}{$\downarrow$ 14.21} & 45.85 \textcolor{green}{$\uparrow$ 8.51}& 47.40 \textcolor{green}{$\uparrow$ 0.75}  \\
\textsc{Gemma-2-9b-IT}& 74.24 \textcolor{green}{$\uparrow$ 1.54} & 31.79 \textcolor{green}{$\uparrow$ 7.36} & 51.76 \textcolor{green}{$\uparrow$ 0.91} & 34.52 \textcolor{green}{$\uparrow$ 0.39} & 48.13 \textcolor{green}{$\uparrow$ 7.87} & 63.76 \textcolor{green}{$\uparrow$ 0.57} & 48.20\textcolor{green}{$\uparrow$ 1.63} \\
\textsc{LLaMa-3.1-8B-IT}& 71.88 \textcolor{green}{$\uparrow$ 4.73} & 34.92 \textcolor{green}{$\uparrow$ 8.01} & 49.50 \textcolor{green}{$\uparrow$ 3.95} & 40.30 \textcolor{green}{$\uparrow$ 1.08} & 52.69 \textcolor{green}{$\uparrow$ 3.91} & 54.85 \textcolor{green}{$\uparrow$ 5.48}&  51.33 \textcolor{green}{$\uparrow$ 4.45} \\ 
\rowcolor{gray!20}
Average & 74.27 \textcolor{green}{$\uparrow$ 2.48} & 	35.07 \textcolor{green}{$\uparrow$ 8.02} & 52.38 \textcolor{green}{$\uparrow$ 2.51} & 37.32 \textcolor{green}{$\uparrow$ 2.60} & 55.31 \textcolor{green}{$\uparrow$ 2.44} & 58.95 \textcolor{green}{$\uparrow$ 4.51} & 56.63\textcolor{green}{$\uparrow$ 2.70}\\
\hline
\end{tabular}
}
\caption{F1-scores results of zero-shot and one-shot evaluation categorized on six impacts and two context length settings. The number in parentheses refers to the improvement with \textcolor{green}{$\uparrow$} or degradation with \textcolor{red}{$\downarrow$} of the evaluation on the mixed-context dataset (1,386 entries) compared to the original dataset (350 entries). The number in the ``Positive Cases'' row denotes the positive label in each impact categorization corresponding to two context-length versions. \textbf{Bold} and \underline{underline} indicate the best and the second-best performance.}
\vspace{-2ex}
\label{tab:f1_comparison}
\end{table*}
\subsection{Evaluated Models}
We evaluate a set of off-the-shelf LLMs on \ourmethod{}. For the multi-label text classification task, we include seven open-source models: \textsc{DeepSeek-V3-671B} \cite{deepseekai2024deepseekv3technicalreport}, \textsc{Llama-3.1-8B-Instruct} \cite{grattafiori2024llama}, \text{Mistral-7B-Instruct}~\cite{jiang2023mistral7b}, \textsc{Mixtral-8x7B-Instruct} \cite{jiang2024mixtralexperts}, \textsc{Mistral-24B-Instruct} \cite{jiang2024mixtralexperts}, \textsc{gemma-2-9b-it} \cite{gemmateam2024gemma2improvingopen}, \textsc{Qwen2.5-7B-Instruct}, and \textsc{Qwen2.5-14B-Instruct} \cite{qwen2025qwen25technicalreport}; and three closed-source models: \textsc{GPT-3.5-Turbo}, \textsc{GPT-4} \cite{openai2024gpt4technicalreport}, and \textsc{GPT-4o} \cite{openai2024gpt4ocard}. For the ranking-based QA task, we evaluate \textsc{GPT-3.5-Turbo}, \textsc{Qwen2.5-7B-Instruct}, \textsc{Qwen2.5-14B-Instruct}, \textsc{Mistral-7B-Instruct}, and \textsc{Llama-3.1-8B-Instruct}. The relatively smaller models (with 7B size) ensure the latency requirements~\cite{sun2023chatgpt}.
The used models for the two tasks cover different sizes and support the input length of at least 8k tokens. The version details are provided in Appendix~\ref{Model_sources}.

\subsection{Implementation Details}
\noindent \textbf{Multi-label Classification.}
The multi-label classification is conducted on each evaluated LLM by the same prompt provided in Appendix~\ref{prompt_temp}.
Different from traditional methods that decompose multi-label text classification into multiple binary classification tasks \cite{BOUTELL20041757, Deep_Text_Classification}, we simultaneously identify all relevant disruptive weather impacts for each input by calling the LLM once.
The example of in-context learning in the one-shot setting is handcrafted with a complex sample detailing multiple impacts. \\

\noindent \textbf{Ranking-based Question-Answering.}
We employ \textsc{GPT-4o} for pseudo question generation with default hyper-parameters.
For ranking evaluation, we adopt the sliding window mechanism within LLM-based ranker implementation following the state-of-the-art study~\cite{sun2023chatgpt} to reduce the potential negative effect of noisy long contexts.
Specifically, each article in the candidate pool was segmented into three chunks, and then the initial ranking was performed independently within each chunk. 
The used prompts for both stages are provided in Appendix~\ref{task2_prompt}.

To ensure stable results, following previous studies~\cite{NEURIPS2023_8b8a7960}, all LLMs were evaluated with the temperature set to $0$, and the reported performance is the average value of running the experiments three times.

\section{Experiments}
\subsection{Results of Multi-label Classification}

\begin{table}[t]
\centering
\resizebox{\columnwidth}{!}{%
\begin{tabular}{lcc}
\hline
\textbf{Model} & \textbf{Zero-Shot} & \textbf{One-Shot} \\ \hline
\textsc{GPT-4o} & \underline{32.28} \textcolor{green}{$\uparrow$ 0.29} & \textbf{31.99} \textcolor{red}{$\downarrow$ 0.85} \\
\textsc{GPT-4} & 22.19 \textcolor{green}{$\uparrow$ 0.38} & 20.46 \textcolor{green}{$\uparrow$ 0.11} \\ 
\textsc{GPT-3.5-turbo} & 21.61 \textcolor{red}{$\downarrow$ 0.18} & 12.39 \textcolor{green}{$\uparrow$ 6.18} \\ 
\hline
\textsc{DeepSeek-V3-671B} & \textbf{34.01} \textcolor{red}{$\downarrow$ 1.72} & \textbf{31.99} \textcolor{red}{$\downarrow$ 0.28} \\ 
\textsc{Mistral-24B-IT} & 19.88 \textcolor{red}{$\downarrow$ 1.02} & \underline{25.65} \textcolor{red}{$\downarrow$ 1.08} \\ 
\textsc{Mixtral-8x7B-IT} & 25.07 \textcolor{red}{$\downarrow$ 0.50} & 19.88 \textcolor{green}{$\uparrow$ 0.12} \\ 
\textsc{Mistral-7B-IT} & 4.90 \textcolor{red}{$\downarrow$ 0.04} & 8.93 \textcolor{red}{$\downarrow$ 3.50} \\ 
\textsc{Qwen2.5-14B-IT} & 19.02 \textcolor{red}{$\downarrow$ 2.45} & 18.16 \textcolor{red}{$\downarrow$ 0.73} \\
\textsc{Qwen2.5-7B-IT} & 27.38 \textcolor{red}{$\downarrow$ 6.52} & \underline{25.65} \textcolor{red}{$\downarrow$ 1.36} \\ 
\textsc{Gemma-2-9B-IT} & 15.56 \textcolor{green}{$\uparrow$ 0.44} & 9.51 \textcolor{red}{$\downarrow$ 1.51} \\ 
\textsc{Llama-3.1-8B-IT} & 12.68 \textcolor{green}{$\uparrow$ 2.18} & 15.56 \textcolor{red}{$\downarrow$ 1.85} \\ 
\rowcolor{gray!20}
Average & 21.96 \textcolor{red}{$\downarrow$ 0.57} & 20.80 \textcolor{red}{$\downarrow$ 0.10} \\ 
\hline
\end{tabular}
}
\caption{Row-wise accuracy performance across different models and prompting strategies. The used notions are the same as Table~\ref{tab:f1_comparison}.}
\vspace{-2ex}
\label{tab:row-wise}
\end{table}

Table~\ref{tab:f1_comparison} and Table~\ref{tab:row-wise} show the performance of the evaluated LLMs on \ourmethod{} for the settings of categorized by six societal impacts with different context lengths, overall row-wise evaluation, and divided into two periods, respectively. The additional experimental results are provided in Appendix~\ref{Additional_exp}.
We have the observations as follows. \\

\noindent\textbf{LLMs struggle to understand disruptive weather impacts.} 
Table \ref{tab:f1_comparison} shows that the F1-score for multi-label classification remains consistently low across models, especially among the political and ecological categories. The financial, agricultural, and human health impacts categories perform slightly better but still exhibit suboptimal results at $55\%$. The low performance might be attributed to the challenges in these categories with abstract and context-dependent narratives. Different from the infrastructure category (with the highest performance), which describes the impacts of weather events explicitly, e.g., ``bridges destroyed'' and ``power outages'', the descriptions in other categories are usually more abstract. For example, the financial damage could be embedded within discussions of damaged infrastructure or agricultural setbacks, which makes it more difficult for models to distinguish them as standalone impacts.

Table~\ref{tab:row-wise} shows row-wise performance, in which the model must identify the given sample correctly for each involved category, the performance of classification drops dramatically due to the more precise requirement.
Thus, a sophisticated model is expected to understand the complex societal effects of historical narratives via reasoning~\cite{wei2022chain,zhang2025entropy,zhang2025ratt}. \\

\noindent\textbf{Long-context LLMs not always be strong on long-context de-noising.} 
The results in Table~\ref{tab:f1_comparison} show that, when the original long-context is segmented into smaller chunks, the classification accuracy increases in most cases. 
These improvements suggest that smaller chunks help models focus on relevant information and thus minimizing distraction from extraneous content.
Even the used models are claimed with long-context capacity, more precise split that reduces potential noise is still effective for context de-noising, which is consistent with previous studies~\cite{sun2024chatgptgoodsearchinvestigating}.

However, we also find that this trend is not observed with the row-wise accuracy evaluation. This is due to the evaluation bias, where the F1-score measures precision and recall per category, and benefits from partial correctness. Row-wise accuracy requires an exact match across all labels. 
The small chunks might be helpful to improve the classification of one of the categories but not enough to correct all labels.
Thus, the helpfulness of long-context de-noising via small chunks is not obvious in overall performance. \\

\noindent\textbf{In-context Learning offers limited improvement.} 
The in-context learning is achieved by providing one demonstration as the one-shot example for model decision.
Compared zero-shot and one-shot performance in Table~\ref{tab:f1_comparison}, we find that providing a single example in the prompt offers limited benefits and might decrease the performance in some cases.
Such a phenomenon implies that the LLMs lack sufficient knowledge to disambiguate disruptive weather impacts even with enhanced examples for knowledge arousing.
These results indicate that our \ourmethod{} is challenging for LLMs to understand disruptive weather impact. \\

\begin{table}[t]
\centering
\resizebox{\columnwidth}{!}{%
\begin{tabular}{lcc}
\hline
~ & \textbf{Historical} & \textbf{Modern} \\ \hline
Cases & 200 (504) & 150 (882)\\
\hline
\textsc{GPT-4o} & \textbf{70.19} \textcolor{green}{$\uparrow$ 1.50} & \textbf{68.59} \textcolor{green}{$\uparrow$ 0.59} \\ 
\textsc{GPT-4} & \underline{65.54} \textcolor{green}{$\uparrow$ 0.66} & 53.81 \textcolor{green}{$\uparrow$ 4.50} \\ 
\textsc{GPT-3.5-turbo} & 57.27 \textcolor{green}{$\uparrow$ 4.78} & 52.21 \textcolor{green}{$\uparrow$ 5.92} \\ 
\hline
\textsc{DeepSeek-V3-671B} & 65.42 \textcolor{green}{$\uparrow$ 3.91} & \underline{64.74} \textcolor{green}{$\uparrow$ 2.05} \\ 
\textsc{Mistral-24B-IT} & 62.30 \textcolor{green}{$\uparrow$ 4.13} & 63.33 \textcolor{red}{$\downarrow$ 1.68} \\ 
\textsc{Mixtral-8x7B-IT} & 58.31 \textcolor{green}{$\uparrow$ 4.28} & 50.11 \textcolor{green}{$\uparrow$ 6.14} \\ 
\textsc{Mistral-7B-IT} & 55.27 \textcolor{green}{$\uparrow$ 1.50} & 49.22 \textcolor{green}{$\uparrow$ 2.78} \\ 
\textsc{Qwen2.5-14B-IT} & 57.75 \textcolor{green}{$\uparrow$ 2.30} & 57.75 \textcolor{red}{$\downarrow$ 5.48} \\ 
\textsc{Qwen2.5-7B-IT} & 54.29 \textcolor{green}{$\uparrow$ 0.66} & 40.85 \textcolor{green}{$\uparrow$ 4.14} \\ 
\textsc{Gemma-2-9B-IT} & 60.41 \textcolor{green}{$\uparrow$ 0.90} & 52.80 \textcolor{green}{$\uparrow$ 1.97} \\ 
\textsc{Llama-3.1-8B-IT} & 55.30 \textcolor{green}{$\uparrow$ 5.08} & 50.67 \textcolor{green}{$\uparrow$ 4.98} \\ 
\rowcolor{gray!20}
Average & 60.19 \textcolor{green}{$\uparrow$ 2.70} & 54.92 \textcolor{green}{$\uparrow$ 2.36} \\ 
\hline
\end{tabular}%
}
\caption{F1-score performance across historical and modern impact categories in zero-shot setting. The used notations are the same as Table~\ref{tab:f1_comparison}.}
\vspace{-2ex}
\label{tab: period_results}
\end{table}

\begin{table}[t!]
\centering
\resizebox{\columnwidth}{!}{%
\begin{tabular}{lcccc}
\hline
\textbf{Model} & \textbf{Hit@1} & \textbf{nDCG@5} & \textbf{Recall@5} & \textbf{MRR} \\
\hline
\textsc{GPT-3.5-Turbo} & \underline{62.09} & 67.31 & 71.04 & 66.90 \\
\textsc{Mistral-7B} & 6.21 & 15.86 & 25.16 & 14.82 \\
\textsc{QWen-2.5-14B} & \textbf{82.09} & \textbf{86.34} & \textbf{85.48} & \textbf{89.55} \\
\textsc{QWen-2.5-7B} & 42.69 & 61.80 & 75.52 & 58.04 \\
\textsc{Llama-3.1-8B} & 64.18 & \underline{70.85} & \underline{75.82} & \underline{69.90} \\
\hline
\multicolumn{5}{c}{Result with Bias for Reference} \\
\hline
\textsc{GPT-4o} & 91.94 & 95.54 &  97.91 & 94.88 \\
\hline
\end{tabular}%
}
\caption{The performance of ranking-based QA tasks across different models. }
\label{tab:llm_metrics}
\vspace{-2ex}
\end{table}

\noindent\textbf{Historical narratives are easier for LLMs to understand.} 
The evaluation of different narratives in terms of historical and modern articles is presented in Table~\ref{tab: period_results}. Surprisingly, the evaluated models perform better on the articles recorded in the historical period. 
The reason might be the structured and formal narrative style used to report disruptive weather events in historical periods, which more explicitly highlights cause-and-effect relationships. 
The observation is revealed by the earlier studies (e.g.,~\citealp{mauch2009natural}), where the historical narratives emphasize empirical observations over interpretations, offering a more immediate and naturalistic account of events.
Though the modern text might dominate within the pre-trained corpus, the language patterns used in historical narrative styles are easier for language models to identify, and thus perform better on classifying disruptive weather impacts. \\

\noindent\textbf{Scaling law might hold for disruptive weather impact understanding.} 
As illustrated in Table ~\ref{tab:f1_comparison} and Table~\ref{tab:row-wise}, larger models usually perform better than smaller ones, which is consistent with the scaling law for LLMs~\cite{kaplan2020scaling}. For example, the largest \textsc{DeepSeek-V3-671B} obtains the best results and \textsc{Mistral-24B-IT} outperforms its 7B version.
Although the model size is unavailable in closed-source models, the open-source models with the feasibility of manipulation can be viable alternatives to adaptively work for domain requirements. 
With proper optimization, the second-best \textsc{DeepSeek-V3-671B} for understanding disruptive weather impact might offer performance close to or on par with \textsc{GPT-4o}.

\subsection{Results of Ranking-based QA}

The performance of each evaluated model for ranking-based QA is reported in Table~\ref{tab:llm_metrics}.
We find that \textsc{QWen-2.5-14B-IT} achieves the best performance in this task. \textsc{Llama-3.1-8B} is slightly better than \textsc{GPT-3.5-Turbo} and \textsc{QWen-2.5-7B-IT}, while the \textsc{Mistral-7B-IT} cannot address the tasks related to disruptive weather context.
Notice that the ranking results would contain bias when the evaluated model is used for question generation (\textsc{GPT-4o} in our cases). This is a common phenomenon~\cite{zhou2023don} and needs to be avoided in benchmarking.

The practical open-retrieval setting, i.e., identifying the relevant articles from a huge database, is left for future studies, which could further facilitate knowledge enhancement in understanding disruptive weather impacts.

\section{Conclusion}
In this study, we propose a disruptive weather impact understanding benchmark, \ourmethod{}, to address the lack of datasets and evaluation frameworks in climate change adaptation.
We first process the corpus from newspaper articles and provide comprehensive instruction for impact annotation with each processed article. 

Then, we design multi-label classification and ranking-based QA tasks to evaluate the LLMs in understanding various defined disruptive weather impacts.
Extensive experiments on \ourmethod{} reveal that the existing LLMs struggle with understanding disruptive weather impacts across different style narratives and context settings. 
Our \ourmethod{} enables the community to evaluate the developed systems on disruptive weather impact understanding, which supports the society to learn from and prepare for the impacts of climate change.

\section*{Limitation}
Although \ourmethod{} provides valuable insights (e.g., exhibit the strengths and weaknesses of various society impact understanding) about evaluating LLMs on disruptive weather, it may have potential biases in underrepresented historical events and linguistic variations. 
Future work could expand the range of evaluated models, strategies, and designed tasks to further strengthen the evaluations.

\section*{Ethical Considerations}
Our primary data source is a corpus of historical digitized newspapers, obtained through collaboration with an official organization, which should be anonymous at this moment. 
This organization preserves the copyright of the newspaper articles and has been granted permission to publish this subset of articles for benchmark build-up to facilitate the research community. 
Thus, the data is publicly available and thus no potential privacy or content safety concerns.
Additionally, topic-aware article selection is conducted by researchers specializing in historical climate analysis to ensure the dataset is not biased on specific time and location.
This research contributes to the broader societal goal of understanding historical disruptive weather impacts to help society defend its vulnerabilities from disasters. 
The interpretation of weather-related disruptions in historical newspapers might be influenced by demographic and contextual factors, which is similar to other text datasets generated through crowd-sourcing with inherent challenges in ensuring that dataset labels are fully representative of diverse societal perspectives \cite{talat-etal-2022-machine}.

\section*{Acknowledgment}
Our primary data source is a corpus of three digitized newspapers (La Presse, La Patrie and Montreal Gazette), obtained through collaboration with the McGill University Library and Archives and the Bibliothèque nationale du Québec.
We would like to thank DRAW McGill for their guidance throughout this project, especially Dr. Victoria Slonosky, whose expertise in historical meteorology was instrumental in accessing the corpus and advising on OCR correction.
We are also deeply grateful for the support provided by OpenAI Grants and RBC Borealis AI.

\bibliography{main}
\clearpage
\appendix
\section*{Appendix}
\section{Post-OCR Correction}
\subsection{Example of OCR-Digitized Text} 
\label{OCR_example}
\begin{figure}[h]  
    \centering  
    \includegraphics[width=0.49\textwidth]{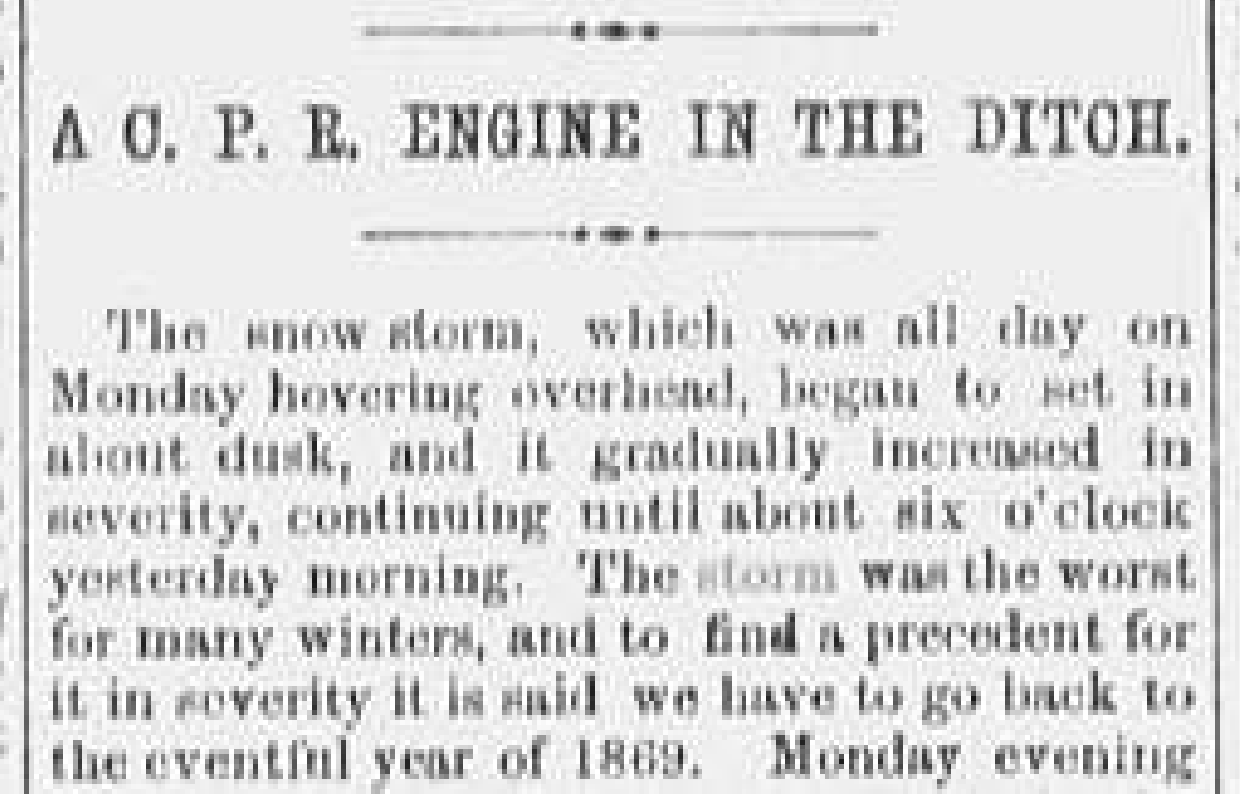}  
    \caption{Example of OCR-digitized text from the \textit{Illustrated Montreal Gazette}, dated February 18, 1885.}  
    \label{fig:ocr_example}  
    \vspace{-5pt}  
\end{figure}

Figure~\ref{fig:ocr_example} presents an example of OCR-digitized text from the \textit{Illustrated Montreal Gazette}, dated February 18, 1885. This excerpt, titled "Snowstorm\_delays," was retrieved from \textit{ProQuest} and illustrates the typical noise and distortions introduced by OCR processing in historical newspaper archives.

\subsection{Post-OCR Correction Instruction} 
\label{OCR_instruct}
To reduce the substantial noise in OCR-digitized text, \textsc{GPT-4o} was used for post-OCR correction to enhance text quality. The specific prompt used for this process is presented in Table~\ref{OCR_guidelines}. 

\begin{table}[h]
\centering
\begin{tcolorbox}[colback=gray!10, colframe=black, width=\columnwidth, title=Post-OCR Correction Instruction]
    \fontsize{10pt}{11pt}\selectfont
    \parbox{\columnwidth}
        \texttt{{
    You are an expert OCR correction assistant specializing in historical newspaper text. Your task is to:
    \begin{enumerate}
        \item Correct OCR errors while preserving the original text’s meaning, structure, and formatting.
        \item Accurately retain proper nouns, dates, numbers, and domain-specific terminology.
        \item Maintain paragraph breaks, section headers, bylines, and other structural elements.
        \item Remove extraneous characters (e.g., unnecessary punctuation, OCR artifacts) without altering the content.
        \item Properly reconstruct hyphenated words that were split across lines.
        \item Standardize spacing by eliminating extra spaces and ensuring a consistent format.
        \item Return the corrected text as a single continuous line, with no newline characters.
    \end{enumerate}
\vspace{10pt}
    \noindent\textbf{NOTE:} Do not include explanations, summaries, or additional comments. Only return the corrected text in the specified format.
        }}
\end{tcolorbox}
\caption{Prompts used for Post-OCR correction.}
\label{OCR_guidelines}
\end{table}

\subsection{Post-OCR Correction Evaluation}
\label{OCR_evaluation}
To assess the effectiveness of our post-OCR correction process, we evaluated GPT-4o’s output against human-annotated corrections on a randomly selected sample of 50 articles. The results demonstrate the high accuracy of the automated corrections:
\begin{table}[h]
\centering
\begin{tabular}{lccc}
\hline
\textbf{Metric} & \textbf{1-gram} & \textbf{2-gram} & \textbf{3-gram / L} \\
\hline
\textbf{BLEU} & 0.9115 & 0.8935 & 0.8773 \\
\textbf{ROUGE} & 0.9476 & 0.9190 & 0.9438 \\
\hline
\end{tabular}
\caption{Evaluation metrics comparing GPT-4o OCR corrections to human annotations (50 samples). ROUGE-L is listed under the third column.}
\end{table}

The consistently high BLEU and ROUGE scores indicate that GPT-4o's corrections closely align with human-edited versions, validating its effectiveness for improving text quality prior to downstream analysis.

\section{Annotation Guidelines for Multi-Label Classification}
\subsection{Definition of Primary Disaster Categories}  
\label{Annotation_guideline}

Using Latent Dirichlet Allocation, the dataset was categorized into 15 primary weather event types. 
The major weather categories are listed in Table \ref{tab:disaster_categories}:  

\begin{table}[htbp]
    \centering
    \resizebox{\columnwidth}{!}{
    \begin{tabular}{|c|c|c|}
        \hline
        \textbf{Disaster Type} & \textbf{Definition} & \textbf{Example} \\ \hline
        \textbf{Blizzard} & {\small Severe snowstorm} & {\small \textit{Whiteout conditions}} \\ \hline
        \textbf{Cold} & {\small Low temperatures} & {\small \textit{Frostbite risk}} \\ \hline
        \textbf{Deluge} & {\small Heavy rainfall} & {\small \textit{Flash flooding}} \\ \hline
        \textbf{Drought} & {\small Prolonged dryness} & {\small \textit{Water scarcity}} \\ \hline
        \textbf{Flood} & {\small Overflowing water} & {\small \textit{River flooding}} \\ \hline
        \textbf{Heat} & {\small High temperatures} & {\small \textit{Heat exhaustion}} \\ \hline
        \textbf{Heatwave} & {\small Extended hot weather} & {\small \textit{Record-breaking heat}} \\ \hline
        \textbf{Ice} & {\small Frozen precipitation} & {\small \textit{Slippery surfaces}} \\ \hline
        \textbf{Rain} & {\small Liquid precipitation} & {\small \textit{Persistent showers}} \\ \hline
        \textbf{Freezing} & {\small Below 0°C conditions} & {\small \textit{Frost formation}} \\ \hline
        \textbf{Snow} & {\small Frozen precipitation} & {\small \textit{Accumulating snowfall}} \\ \hline
        \textbf{Snowstorm} & {\small Intense snowfall} & {\small \textit{Reduced visibility}} \\ \hline
        \textbf{Storm} & {\small Severe weather event} & {\small \textit{Strong winds/rain}} \\ \hline
        \textbf{Thunder} & {\small Sound from lightning} & {\small \textit{Loud rumbling}} \\ \hline
        \textbf{Torrential} & {\small Extreme rainfall} & {\small \textit{Flooding risk}} \\ \hline
    \end{tabular}
    }
    \caption{Primary weather types and their definitions.}
    \label{tab:disaster_categories}
\end{table}

\subsection{Background}  
In the absence of standardized impact records (e.g., flood-related property damage, injuries due to ice accumulation, power outages, and road closures), we assessed vulnerabilities and resilience based on the consequences of weather events and how they have changed since the 19th century. To do so, we categorized disruptive weather impacts into six primary groups — Infrastructural, Agricultural, Ecological, Financial, Human Health, and Political — following previous studies ~\cite{imran_weather}.

To ensure high-quality and consistent annotations, the task was conducted using a set of specific instructions reviewed by meteorological experts. The annotation guideline and the categories definition are provided in Table \ref{annotatorinstruct} and Table \ref{tab:impact_categories}, respectively.

Notably, the same instruction guidance is contained within the prompts for LLMs in Appendix~\ref{Annotation_instruction_whole} to perform impact classification, following a binary output approach for each category.

\subsubsection{Multi-Impact Labeling Format}  

Annotators are tasked with determining whether an article includes descriptions that correspond to the impact categories defined in Table \ref{tab:impact_categories}. Each article is assigned a label based on the presence or absence of relevant descriptions.

\begin{itemize}
    \item \textbf{1} – At least one mention of the relevant topic is identified.
    \item \textbf{0} – No relevant description is identified.
\end{itemize}  

\paragraph{Special Case}  
When an article describes multiple types of impact, each mentioned impact category is labelled as "1".

\begin{table}[h]
\centering
\begin{tcolorbox}[colback=gray!10, colframe=black, width=\linewidth, title=A Special Case Example, label=]
    \fontsize{11pt}{11pt}\selectfont
    \parbox{\textwidth}
    
        \textbf{Input Text}: "Severe Storm Wreaks Havoc on Coastline. Bayport, September 15. A violent tempest swept the eastern seaboard last night, leaving a trail of devastation in its wake. The cargo vessel Harbor Star collided with a fishing trawler in the churning waters, capsizing the smaller craft and claiming three lives. Fierce winds reduced docks and piers to splinters, bringing commercial shipping to a standstill. The storm’s toll is estimated to exceed \$200,000, with Bayport Textile Mills filing for financial restructuring, placing 150 jobs in jeopardy. Hospitals are overwhelmed with storm-related injuries and illnesses, and emergency shelters are strained beyond capacity. The community now faces the arduous task of recovery."\\
        \textbf{Output}: "Infrastructural: true, Agricultural: false, Ecological: false, Financial: true, Human Health: true, Political: false"
        
\end{tcolorbox}
\caption{A special case with multiple positive labels and is used for one-shot learning.}
\label{one-shot-ex}
\end{table}

\subsubsection*{Dataset Statistics and Article Topics}
\label{appendix:topics}
To provide additional transparency regarding the dataset used in our analysis, we include detailed statistics of the included articles. The average number of tokens per article is 2987.4 in long-context settings and 781.3 in mixed-context settings.

Furthermore, we categorized the articles based on their associated weather-related topics. The table below shows the distribution of topics across the corpus:

\begin{table}[h]
\centering
\begin{tabular}{ll|ll}
\toprule
\textbf{Weather} & \textbf{Count} & \textbf{Weather} & \textbf{Count} \\
\midrule
Snowstorm & 34 & Flood & 22 \\
Rain      & 32 & Cold & 21 \\
Drought   & 31 & Snow & 17 \\
Ice       & 30 & Torrential & 16 \\
Storm     & 30 & Blizzard & 15 \\
Thunder   & 29 & Heatwave & 13 \\
Deluge    & 28 & Heat & 4 \\
Freezing  & 28 & & \\
\bottomrule
\end{tabular}
\caption{Distribution of weather-related topics in the dataset. Each article may be assigned one or more topics based on content.}
\label{tab:topic_distribution}
\end{table}

\subsubsection{Instructions for Annotators} 
\label{Annotation_Instruction}
The instructions for annotators are provided in Table \ref{annotatorinstruct}.
The annotation process was conducted by members of a research group specializing in uncovering the history of a region's climate change through the regional historical weather records.
Their expertise can ensure the accuracy and reliability of annotations.

\section{Instructions}
\label{Annotation_instruction_whole}
\subsection{Multi-Label Classification Instructions}
The Multi-Label Classification instructions template in Table \ref{category_annot} is designed for both zero-shot and one-shot classification tasks.  
\begin{itemize}
    \item Zero-Shot: The model is given only the classification instructions and the input text.
    \item One-Shot for In-Context Learning: The model is provided with a demonstration for predicting a new sample. One example of demonstration we used is shown in Table \ref{one-shot-ex}.
\end{itemize}

\subsection{Prompt Template for Multi-Label Text Classification}
\label{prompt_temp}
Table \ref{category_annot} presents the prompt designed to analyze historical newspaper texts and classify them into six distinct impact categories based on explicit mentions of weather-related events. The prompt is structured in alignment with the definitions provided in Table \ref{tab:impact_categories}, which details the scope of each impact category, including Infrastructural, Agricultural, Ecological, Financial, Human Health, and Political impacts. The classification task is binary (true/false), requiring the model to identify whether the text explicitly mentions any of the defined impacts. The guidelines emphasize focusing on direct and immediate effects, ensuring that classifications are based solely on explicit references within the text. This prompt was used to evaluate multi-label classification models.

\begin{table*}[h!]
\centering
\resizebox{\textwidth}{!}{%
\begin{tabular}{lccccccc}
\hline
\textbf{Model} & \textbf{Infrastructural} & \textbf{Political} & \textbf{Financial} & \textbf{Ecological} & \textbf{Agricultural} & \textbf{Human Health} & \textbf{Average} \\ 
\hline
\multicolumn{8}{c}{\textbf{Zero-Shot Performance}} \\
\hline
\textsc{GPT-4o} & \underline{78.96} \textcolor{green}{$\uparrow$ 0.18} & \underline{84.44} \textcolor{red}{$\downarrow$ 0.44} & \underline{76.66} \textcolor{red}{$\downarrow$ 0.95} & 78.39 \textcolor{green}{$\uparrow$ 0.47} & \textbf{84.44} \textcolor{green}{$\uparrow$ 0.42} & \textbf{80.40} \textcolor{red}{$\downarrow$ 0.11} & \underline{80.55} \textcolor{red}{$\downarrow$ 0.07} \\
\textsc{GPT-4} & 72.33 \textcolor{green}{$\uparrow$ 2.24} & 76.37 \textcolor{red}{$\downarrow$ 2.37} & 59.65 \textcolor{green}{$\uparrow$ 0.06} & 80.12 \textcolor{red}{$\downarrow$ 3.26} & 79.25 \textcolor{green}{$\uparrow$ 1.32} & 71.18 \textcolor{green}{$\uparrow$ 0.82} & 73.15 \textcolor{green}{$\uparrow$ 0.17} \\ 
\textsc{GPT-3.5-turbo} & 77.52 \textcolor{green}{$\uparrow$ 3.05} & 78.96 \textcolor{red}{$\downarrow$ 0.67} & 69.45 \textcolor{red}{$\downarrow$ 1.45} & 78.96 \textcolor{red}{$\downarrow$ 1.82} & 80.69 \textcolor{green}{$\uparrow$ 0.74} & 69.16 \textcolor{green}{$\uparrow$ 1.70} & 75.79 \textcolor{green}{$\uparrow$ 0.25} \\ 
\hline
\textsc{DeepSeek-V3-671B} & \textbf{80.98} \textcolor{green}{$\uparrow$ 0.45} & \textbf{85.59} \textcolor{green}{$\uparrow$ 1.84} & \textbf{77.81} \textcolor{green}{$\uparrow$ 0.48} & \underline{81.56} \textcolor{red}{$\downarrow$ 2.13} & \underline{83.57} \textcolor{green}{$\uparrow$ 1.00} & \underline{77.23} \textcolor{green}{$\uparrow$ 1.06} &\textbf{ 81.12} \textcolor{green}{$\uparrow$ 0.45} \\ 
\textsc{Mistral-24B-IT} & 76.30 \textcolor{red}{$\downarrow$ 0.59} & 74.35 \textcolor{red}{$\downarrow$ 3.78} & 69.45 \textcolor{red}{$\downarrow$ 1.16} & 76.66 \textcolor{green}{$\uparrow$ 0.20} & 82.42 \textcolor{green}{$\uparrow$ 0.44} & 69.45 \textcolor{green}{$\uparrow$ 0.26} & 74.77 \textcolor{red}{$\downarrow$ 0.77} \\ 
\textsc{Mixtral-8x7B-IT} & 75.36 \textcolor{green}{$\uparrow$ 2.16} & 80.29 \textcolor{red}{$\downarrow$ 2.42} & 69.86 \textcolor{red}{$\downarrow$ 4.06} & \textbf{84.93} \textcolor{red}{$\downarrow$ 3.32} & 78.26 \textcolor{green}{$\uparrow$ 3.92} & 68.12 \textcolor{green}{$\uparrow$ 1.99} & 76.14 \textcolor{green}{$\uparrow$ 0.05} \\ 
\textsc{Mistral-7B-IT} & 77.23 \textcolor{green}{$\uparrow$ 2.48} & 50.14 \textcolor{red}{$\downarrow$ 1.00} & 34.58 \textcolor{red}{$\downarrow$ 1.15} & 74.06 \textcolor{red}{$\downarrow$ 3.49} & 72.05 \textcolor{green}{$\uparrow$ 0.52} & 75.22 \textcolor{green}{$\uparrow$ 0.49} & 63.88 \textcolor{green}{$\uparrow$ 0.47} \\ 
\textsc{Qwen2.5-14B-IT} & 73.20 \textcolor{red}{$\downarrow$ 2.91} & 70.03 \textcolor{red}{$\downarrow$ 3.46} & 66.28 \textcolor{red}{$\downarrow$ 2.28} & 66.86 \textcolor{red}{$\downarrow$ 6.29} & 75.79 \textcolor{red}{$\downarrow$ 7.22} & 70.32 \textcolor{red}{$\downarrow$ 4.03} & 70.41 \textcolor{red}{$\downarrow$ 4.37} \\ 
\textsc{Qwen2.5-7B-IT} & 72.05 \textcolor{green}{$\uparrow$ 1.66} & 73.49 \textcolor{red}{$\downarrow$ 8.35} & 67.72 \textcolor{red}{$\downarrow$ 6.58} & 74.35 \textcolor{red}{$\downarrow$ 10.06} & 73.49 \textcolor{red}{$\downarrow$ 5.20} & 64.55 \textcolor{red}{$\downarrow$ 2.84} & 70.94 \textcolor{red}{$\downarrow$ 5.23} \\ 
\textsc{Gemma-2-9b-IT} & 74.00 \textcolor{green}{$\uparrow$ 3.49} & 61.14 \textcolor{red}{$\downarrow$ 6.00} & 54.86 \textcolor{green}{$\uparrow$ 2.26} & 66.29 \textcolor{red}{$\downarrow$ 3.57} & 69.43 \textcolor{red}{$\downarrow$ 5.43} & 64.00 \textcolor{red}{$\downarrow$ 4.00} & 65.37 \textcolor{green}{$\uparrow$ 0.47}\\ 
\textsc{LLaMa-3.1-8B-IT} & 73.49 \textcolor{green}{$\uparrow$ 5.94} & 55.91 \textcolor{green}{$\uparrow$ 0.66} & 62.25 \textcolor{red}{$\downarrow$ 3.96} & 77.52 \textcolor{red}{$\downarrow$ 2.66} & 77.81 \textcolor{green}{$\uparrow$ 1.33} & 68.59 \textcolor{green}{$\uparrow$ 4.55} & 69.26 \textcolor{green}{$\uparrow$ 1.14} \\ 
\rowcolor{gray!20}
Average & 75.58 \textcolor{green}{$\uparrow$ 1.65} & 71.88 \textcolor{red}{$\downarrow$ 2.36} & 64.42 \textcolor{red}{$\downarrow$ 1.71} & 76.34 \textcolor{red}{$\downarrow$ 3.27} & 77.93 \textcolor{red}{$\downarrow$ 0.74} & 70.75 \textcolor{green}{$\uparrow$ 0.00} & 72.82 \textcolor{red}{$\downarrow$ 0.71} \\

\hline
\multicolumn{8}{c}{\textbf{In-Context Learning (One-shot) Performance}} \\
\hline
\textsc{GPT-4o} & \underline{79.25} \textcolor{red}{$\downarrow$ 0.82} & \underline{84.73} \textcolor{red}{$\downarrow$ 1.02} & \underline{74.64} \textcolor{green}{$\uparrow$ 0.07} & \textbf{79.83} \textcolor{red}{$\downarrow$ 1.26} & \textbf{85.01} \textcolor{red}{$\downarrow$ 0.15} & \textbf{80.12} \textcolor{green}{$\uparrow$ 1.02} & \underline{80.60} \textcolor{red}{$\downarrow$ 0.36} \\ 
\textsc{GPT-4} & 70.89 \textcolor{green}{$\uparrow$ 2.82} & 70.61 \textcolor{red}{$\downarrow$ 0.90} & 61.10 \textcolor{red}{$\downarrow$ 1.10} & \underline{79.54} \textcolor{red}{$\downarrow$ 1.25} & 79.25 \textcolor{green}{$\uparrow$ 1.32} & 69.16 \textcolor{green}{$\uparrow$ 2.27} & 71.76 \textcolor{green}{$\uparrow$ 0.52} \\ 
\textsc{GPT-3.5-turbo} & 73.49 \textcolor{green}{$\uparrow$ 3.65} & 73.78 \textcolor{green}{$\uparrow$ 4.51} & 55.33 \textcolor{green}{$\uparrow$ 5.53} & 61.96 \textcolor{green}{$\uparrow$ 9.47} & 77.23 \textcolor{green}{$\uparrow$ 2.77} & 72.33 \textcolor{green}{$\uparrow$ 0.24} & 69.02 \textcolor{green}{$\uparrow$ 4.36} \\ 
\hline
\textsc{DeepSeek-V3-671B} &\textbf{ 80.40} \textcolor{green}{$\uparrow$ 1.03} &\textbf{ 85.88} \textcolor{red}{$\downarrow$ 0.71} & \textbf{78.67} \textcolor{red}{$\downarrow$ 0.67} & 79.83 \textcolor{red}{$\downarrow$ 2.40} & \underline{83.29} \textcolor{green}{$\uparrow$ 1.28} & \underline{77.81} \textcolor{red}{$\downarrow$ 0.38} & \textbf{80.98} \textcolor{red}{$\downarrow$ 0.08}\\ 
\textsc{Mistral-24B-IT} & 76.88 \textcolor{red}{$\downarrow$ 0.02} & 81.21 \textcolor{red}{$\downarrow$ 0.64} & 76.01 \textcolor{red}{$\downarrow$ 2.87} & 78.61 \textcolor{red}{$\downarrow$ 1.18} & 80.06 \textcolor{green}{$\uparrow$ 1.94} & 71.10 \textcolor{red}{$\downarrow$ 1.39} & 77.31 \textcolor{red}{$\downarrow$ 0.69} \\ 
\textsc{Mixtral-8x7B-IT} & 70.59 \textcolor{green}{$\uparrow$ 4.12} & 77.49 \textcolor{red}{$\downarrow$ 1.90} & 62.70 \textcolor{green}{$\uparrow$ 3.77} & 71.38 \textcolor{green}{$\uparrow$ 6.97} & 74.43 \textcolor{green}{$\uparrow$ 1.75} & 65.81 \textcolor{green}{$\uparrow$ 7.43} & 70.40 \textcolor{green}{$\uparrow$ 3.52} \\ 
\textsc{Mistral-7B-IT} & 73.70 \textcolor{green}{$\uparrow$ 1.73} & 69.08 \textcolor{green}{$\uparrow$ 1.21} & 44.22 \textcolor{red}{$\downarrow$ 1.36} & 71.68 \textcolor{red}{$\downarrow$ 4.82} & 67.92 \textcolor{red}{$\downarrow$ 8.78} & 70.52 \textcolor{red}{$\downarrow$ 2.23} & 66.19 \textcolor{red}{$\downarrow$ 2.38} \\ 
\textsc{Qwen2.5-14B-IT} & 76.08 \textcolor{red}{$\downarrow$ 0.37} & 76.66 \textcolor{red}{$\downarrow$ 4.37} & 71.76 \textcolor{red}{$\downarrow$ 0.33} & 68.01 \textcolor{red}{$\downarrow$ 3.72} & 78.96 \textcolor{green}{$\uparrow$ 1.61} & 72.62 \textcolor{green}{$\uparrow$ 1.09} & 74.02 \textcolor{red}{$\downarrow$ 1.18} \\ 
\textsc{Qwen2.5-7B-IT} & 69.45 \textcolor{green}{$\uparrow$ 0.84} & 82.42 \textcolor{red}{$\downarrow$ 0.71} & 63.11 \textcolor{red}{$\downarrow$ 5.68} & 62.82 \textcolor{red}{$\downarrow$ 8.02} & 81.56 \textcolor{red}{$\downarrow$ 3.27} & 60.52 \textcolor{green}{$\uparrow$ 0.62} & 69.81 \textcolor{red}{$\downarrow$ 2.48} \\ 
\textsc{Gemma-2-9b-IT} & 71.14 \textcolor{green}{$\uparrow$ 0.62} & 67.14 \textcolor{red}{$\downarrow$ 3.09} & 50.29 \textcolor{red}{$\downarrow$ 2.31} & 60.57 \textcolor{red}{$\downarrow$ 2.15} & 74.86 \textcolor{green}{$\uparrow$ 2.89} & 68.00 \textcolor{red}{$\downarrow$ 0.79} & 65.33 \textcolor{green}{$\uparrow$ 0.62} \\ 
\textsc{LLaMa-3.1-8B-IT} & 74.06 \textcolor{green}{$\uparrow$ 3.08} & 64.55 \textcolor{green}{$\uparrow$ 2.02} & 55.91 \textcolor{green}{$\uparrow$ 0.38} & 76.95 \textcolor{red}{$\downarrow$ 6.09} & 74.64 \textcolor{red}{$\downarrow$ 0.93} & 69.16 \textcolor{green}{$\uparrow$ 0.55} & 69.05 \textcolor{green}{$\uparrow$ 0.19} \\ 
\rowcolor{gray!20}
Average & 74.17 \textcolor{green}{$\uparrow$ 1.52} & 75.78 \textcolor{red}{$\downarrow$ 0.51} & 63.07 \textcolor{red}{$\downarrow$ 0.42} & 71.93 \textcolor{red}{$\downarrow$ 1.31} & 77.93 \textcolor{green}{$\uparrow$ 0.04} & 70.65 \textcolor{green}{$\uparrow$ 0.77} & 72.26 \textcolor{green}{$\uparrow$ 0.19} \\

\hline
\end{tabular}%
}
\caption{Accuracy by percentage results of zero-shot and one-shot evaluation categorized on six impacts and two context length settings. The
used notations are the same as Table \ref{tab:f1_comparison}.}
\label{additional_model_result}
\end{table*}

\subsection{Prompt Template for Question Answering Ranking}
\label{task2_prompt}
The ranking-based QA task consists of two key components: question generation~\cite{mo2023convgqr} and candidate ranking~\cite{meng2024ranked}.
The prompts used for both stages are provided in Table \ref{rank_gen_prompt} and Table \ref{rank_qa_prompt}, respectively.

\begin{table}[h]
\centering
\begin{tcolorbox}[colback=gray!10, colframe=black, title=Question Generation Template]

    \fontsize{11pt}{11pt}\selectfont
    
    \begin{flushleft}
        \texttt{Given the following passage about \{row['Weather']\}, generate a single, focused question that meets these criteria:}  \\
        \texttt{1. Can be answered using ONLY the information in this passage} \\ 
        \texttt{2. Focuses on the \{impact\_str\} impacts mentioned}  \\
        \texttt{3. Is detailed and specific to this exact situation}  \\
        \texttt{4. Requires understanding the passage's unique context}  \\
        \texttt{5. Cannot be answered by other similar passages about \{row['Weather']\}}  \\
        
        \texttt{Passage:}  
        \texttt{\{row['Text']\}}  
    \end{flushleft}
\end{tcolorbox}
\caption{Instruction used to generate Questions in the ranking-based QA task.}
\label{rank_gen_prompt}
\end{table}

\begin{table}[h]
\centering
\begin{tcolorbox}[colback=gray!10, colframe=black,title=QA Ranking Task Prompt]
    \fontsize{11pt}{11pt}\selectfont
    \begin{flushleft}
        \texttt{\{\\
        "role": "system",\\
        "content": "You are an expert that ranks passages based on their relevance to a given query.\\
        \ \ The most relevant passage should answer the query"\\
        \},\{\\
        "role": "user",\\
        "content": f"Query: \{query\} Rank the following passages [\{start\_idx+1\} to \{start\_idx+len(passages)\}] by relevance."\\
        \}\\
        }

        \texttt{for i, passage in enumerate(passages):\\
        messages.extend([\\
        {"role": "user", "content": f"[{start\_idx+i+1}] {passage}"},\\
        {"role": "assistant", "content": f"Received passage [{start\_idx+i+1}]"}\\
        ])\\
        }

        \texttt{\{\\
        "role": "user",\\
        "content": "Provide ranking as numbers separated by '>',\\
        \ \ e.g., [3] > [1] > [2] > [5] > [4]. No explanation needed."\\
        \}}
    \end{flushleft}
\end{tcolorbox}
\caption{Instruction used in Ranking-based QA task.}
\label{rank_qa_prompt}
\end{table}

\section{Dataset Statistic}
\label{token-length}
\begin{figure}[ht!]
    \centering
    \begin{subfigure}[b]{\columnwidth}
        \includegraphics[width=\columnwidth]{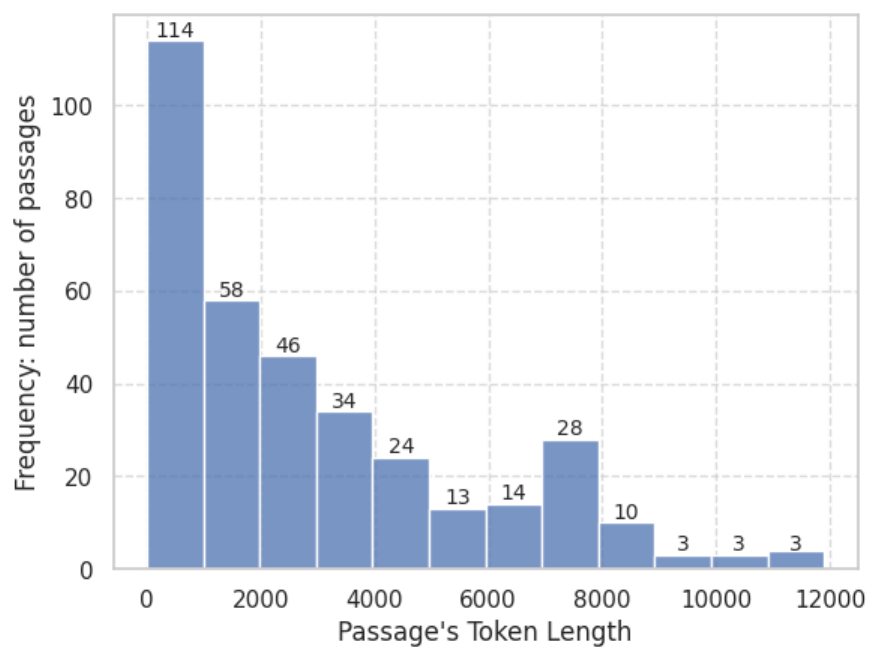}
        \caption{Long context dataset}
        \label{long-context-token}
    \end{subfigure}
    \begin{subfigure}[b]{\columnwidth}
        \includegraphics[width=\columnwidth]{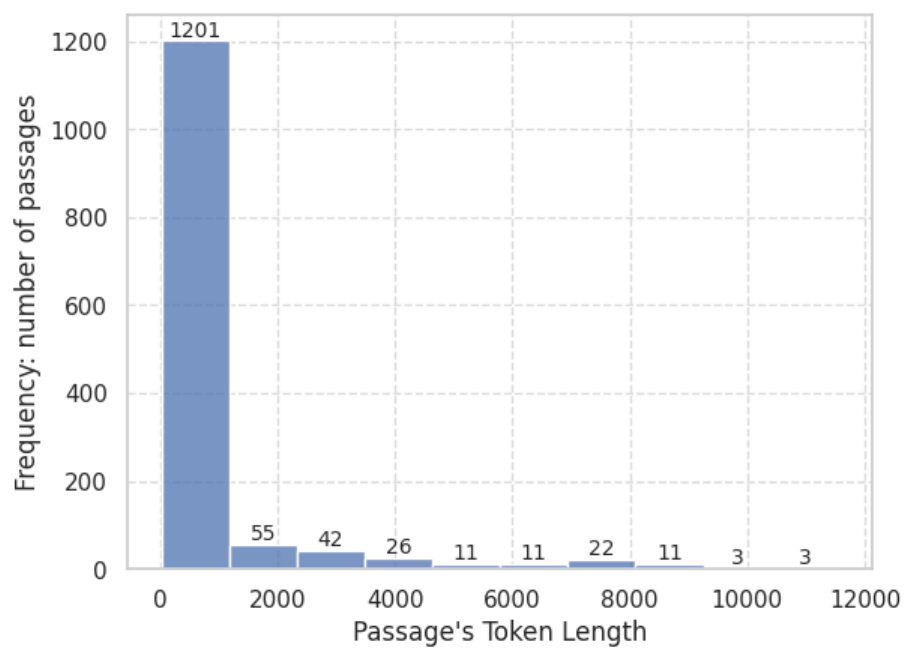} 
        \caption{Mixed context dataset}
        \label{mixed-context-token}
    \end{subfigure}
    \caption{The histogram shows the token length distribution (measured using the \textsc{GPT-4} tokenizer) of articles in the two versions of our dataset.}
    \label{both-token}
\end{figure}

Figure \ref{both-token} presents the token length distribution of passages in two versions of our dataset: (a) the Long Context dataset and (b) the Mixed Context dataset used for context-denoising evaluation. 

The Long Context dataset (Figure \ref{long-context-token}), which contains 350 articles, exhibits a broader distribution of passage lengths, with a significant portion exceeding 2000 tokens. While most passages are concentrated in the lower ranges, a noticeable number extend beyond 8000 tokens, demonstrating the dataset's emphasis on long-form content.

In contrast, the Mixed Context dataset (Figure \ref{mixed-context-token}), which contains 1,386 articles, is heavily skewed toward shorter passages, with an overwhelming majority containing fewer than 2000 tokens. Only a small fraction of passages exceed 4000 tokens, highlighting the dataset's mixed nature, where shorter contexts are predominant.

\begin{table}[t]
\centering
\resizebox{\columnwidth}{!}{%
\begin{tabular}{lcc}

\hline
\textbf{Model} & \textbf{Historical} & \textbf{Modern} \\ \hline
Cases & 200 (504) & 150 (882)\\
\hline
\textsc{GPT-4o} & \textbf{70.24 }\textcolor{green}{$\uparrow$ 1.09}& \textbf{69.05} \textcolor{green}{$\uparrow$ 0.97} \\ 
\textsc{GPT-4} & 62.77 \textcolor{green}{$\uparrow$ 1.81} & 50.89 \textcolor{green}{$\uparrow$ 5.85} \\ 
\textsc{GPT-3.5-turbo} & 60.26 \textcolor{green}{$\uparrow$ 1.18} & 52.63 \textcolor{green}{$\uparrow$ 3.91} \\ 
\hline
\textsc{DEEPSEEK-V3-671B} & \underline{65.77} \textcolor{green}{$\uparrow$ 3.68} & \underline{67.83} \textcolor{green}{$\uparrow$ 0.68} \\ 
\textsc{Mistral-24B-IT} & 63.51 \textcolor{green}{$\uparrow$ 4.16} & 61.27 \textcolor{red}{$\downarrow$ 0.64} \\ 
\textsc{Mixtral-8x7B-IT} & 49.17 \textcolor{green}{$\uparrow$ 8.83} & 41.99 \textcolor{green}{$\uparrow$ 9.01} \\ 
\textsc{Mistral-7B-IT} & 53.11 \textcolor{green}{$\uparrow$ 3.75} & 48.16 \textcolor{red}{$\downarrow$ 0.92} \\ 
\textsc{Qwen-2.5-14B-IT} & 60.66 \textcolor{green}{$\uparrow$ 2.42} & 59.83 \textcolor{green}{$\uparrow$ 0.84} \\ 
\textsc{Qwen2.5-7B-IT} & 52.74 \textcolor{green}{$\uparrow$ 2.60} & 48.84 \textcolor{red}{$\downarrow$ 1.76} \\ 
\textsc{Gemma-2-9B-IT} & 57.82 \textcolor{green}{$\uparrow$ 0.86} & 51.82 \textcolor{green}{$\uparrow$ 2.73} \\ 
\textsc{Llama-3.1-8B-IT} & 55.06 \textcolor{green}{$\uparrow$ 4.89} & 50.75 \textcolor{green}{$\uparrow$ 4.03} \\ 
\rowcolor{gray!20}
Average & 59.19 \textcolor{green}{$\uparrow$ 3.2} & 54.82 \textcolor{green}{$\uparrow$ 2.25} \\ 
\hline
\end{tabular}%
}
\caption{F1-score performance across historical and
modern impact categories in the one-shot setting. The used notations are the same as Table~\ref{tab:f1_comparison}.}
\label{miceof1_oneshot}
\end{table}

\section{The Source of the Models}
\label{Model_sources}
The models evaluated in this paper can be found as follows
\begin{enumerate}
    \item \textsc{GPT-4o}, \textsc{GPT-4} and \textsc{GPT-3.5-turbo} are provided by OpenAI, the base model API document: \url{https://platform.openai.com/docs/models}
    \item \textsc{DEEPSEEK-V3-671B} is upgraded the \textsc{deepseek-chat}, the base model API documents: \url{https://api-docs.deepseek.com/}
    \item \textsc{MIXTRAL-8X7B-IT}\footnote{\url{https://huggingface.co/mistralai/Mixtral-8x7B-Instruct-v0.1}}, \textsc{MISTRAL-24B-IT}\footnote{\url{https://huggingface.co/mistralai/Mistral-Small-24B-Instruct-2501}}, \textsc{MISTRAL-7B-IT}\footnote{\url{https://huggingface.co/mistralai/Mistral-7B-Instruct-v0.2}}, \textsc{LLAMA-3.1-8B-IT}\footnote{\url{https://huggingface.co/meta-llama/Llama-3.1-8B-Instruct}}, \textsc{QWEN2.5-14B-IT}\footnote{\url{https://huggingface.co/Qwen/Qwen2.5-14B-Instruct}}, \textsc{QWEN2.5-7B-IT}\footnote{\url{https://huggingface.co/Qwen/Qwen2.5-7B-Instruct}} and \textsc{GEMMA-2-9B-IT}\footnote{\url{https://huggingface.co/google/gemma-2-9b-it}}, are base models weights from the Huggingface website: \url{https://huggingface.co/}
\end{enumerate}

\section{Computation Cost}
For the large proprietary models (e.g., GPT-4o), conducting a one-time evaluation on our WXImpactBench costs approximately \$3 for multi-label classification tasks and \$5.5 for ranking-based QA tasks. For all open-source models, evaluations were performed on a system with two NVIDIA A6000 (32GB) GPUs. The relatively modest computational requirements highlight the accessibility of our benchmark for researchers with limited computational resources, while still enabling comprehensive evaluation of state-of-the-art models

\section{Additional Experimental Results}
\label{Additional_exp}
LLMs might be more effective in historical narratives. Table \ref{miceof1_oneshot} presents the performance of the evaluated LLMs on \ourmethod{} across historical and modern impact categories in the one-shot setting. The results are categorized based on six societal impact dimensions with varying context lengths. 

\section{Annotation Guidelines}
\label{humman_annot_guide}
To ensure a high-quality evaluation of historical weather impact analysis, we developed a structured annotation framework for meteorology experts. The goal of this annotation is to create a reliable benchmark for assessing the ability of LLMs to understand and classify disruptive weather-related societal and environmental impacts.
The detailed annotation guidelines are provided in Table \ref{annotatorinstruct}, outlining the task objectives, category definitions, and better practices for identifying and classifying weather impacts in historical texts. 

\begin{table}[!h]
\centering
\begin{tcolorbox}[colback=gray!10, colframe=black, width=0.5\textwidth, title=Instructions For Annotators]
    \fontsize{10pt}{11pt}\selectfont
    \textbf{Annotation Guidelines for Historical Weather Impact Analysis}\\[3pt]
    This document provides comprehensive guidelines for annotators who analyze historical newspaper articles describing disruptive weather events. The primary objective is to identify and categorize six distinct impact categories within each text. This analysis will facilitate comparisons with the performance of large language models in understanding weather-related impacts across various societal and environmental dimensions.\\[3pt]
    \textbf{Task Overview}\\
    Annotators will examine historical newspaper articles documenting disruptive weather events. The analysis requires the identification of impacts across six categories: infrastructural, agricultural, ecological, financial, human health, and political. 
    Please refer to Table \ref{tab:impact_categories} for the definitions of these categories. \\[3pt]
    \textbf{Note}: While specific examples are provided for each impact category, annotators should apply their meteorological expertise to identify and classify impacts beyond these examples, maintaining a comprehensive analytical approach.\\[3pt]
    If you have any questions, please feel free to contact us. Thank you for your invaluable support!
\end{tcolorbox}
\caption{Instructions for annotators}
\label{annotatorinstruct}
\end{table}

\begin{table*}[!h]
    \centering
    \renewcommand{\arraystretch}{1.2}
    \begin{tabular}{|p{3.5cm}|p{11cm}|}
        \hline
        \textbf{Category} & \textbf{Definition} \\ 
        \hline
        
        \textbf{Infrastructural \newline Impact} &  
        Examines weather-related damage or disruption to physical infrastructure and essential services. Includes structural damage to buildings, roads, and bridges; disruptions to transportation (e.g., railway cancellations, road closures); interruptions to utilities (e.g., power, water supply); failures in communication networks; and industrial facility damage. Both immediate physical damage and service disruptions should be considered. \\  
        \hline
        
        \textbf{Agricultural Impact} &  
        Focuses on weather-related effects on farming and livestock management. Includes crop yield variations; direct damage to crops, timber, or livestock; modifications to farming schedules; disruptions to food production and supply chains; impacts on farming equipment; and changes in agricultural inputs (e.g., soil conditions, water availability, fertilizers, animal feed). Both immediate and long-term effects should be considered. \\  
        \hline
        
        \textbf{Ecological Impact} &  
        
        Examines effects on natural environments and ecosystems. Includes changes in biodiversity; impacts on wildlife populations and behavior; effects on non-agricultural plant life; habitat modifications (e.g., forests, wetlands, water bodies); changes in hydrological systems (e.g., river levels, lake conditions); and urban plant life impact. Immediate environmental changes should be prioritized. \\  
        \hline
        
        \textbf{Financial Impact} &  
        
        Analyzes economic consequences of weather events. Includes direct monetary losses; business disruptions requiring financial intervention; market fluctuations; impacts on tourism and local economies; and insurance claims or economic relief measures. The focus should be on explicit financial impacts rather than inferred consequences. \\  
        \hline
        
        \textbf{Human Health\newline Impact} &  
        
        Examines both physical and mental health effects. Includes direct injuries or fatalities (including cases where one or more casualties are explicitly mentioned); increased risks of weather-related illnesses; mental health consequences (e.g., stress, anxiety); impacts on healthcare accessibility; and long-term health implications. Both short-term and chronic health effects should be considered. \\  
        \hline
        
        \textbf{Political Impact} &  
        
        Evaluates governmental and policy responses to weather events. Includes government decision-making and policy changes; shifts in public opinion or political discourse; effects on electoral processes; international aid and relations; and debates on disaster preparedness and response. Both direct political reactions and policy implications should be analyzed. \\  
        \hline
    \end{tabular}
    \caption{Impact categories and their definitions}
    \label{tab:impact_categories}
\end{table*}

\begin{table*}[!h]
\centering
\begin{tcolorbox}[colback=gray!10, colframe=black, sharp corners=southwest, width=\textwidth, title=Multi-Label Classification Task: Zero-Shot Instruction Template]
    \fontsize{10pt}{11pt}\selectfont
    \parbox{\textwidth}
        \texttt{{
        Given the following historical newspaper text:\\
        "\{instruction\}"\\

        Analyze the text and provide a binary classification (respond ONLY with 'true' or 'false') for each impact category based on explicit mentions in the text. Follow these specific guidelines\:\\
        
        1. \textbf{Infrastructural Impact}: Classify as 'true' if the text mentions any damage or disruption to physical infrastructure and essential services. This includes structural damage to buildings, roads, or bridges; any disruptions to transportation systems such as railway cancellations or road closures; interruptions to public utilities including power and water supply; any failures in communication networks; or damage to industrial facilities. Consider only explicit mentions of physical damage or service disruptions in your classification.\\
        
        2. \textbf{Agricultural Impact}: Classify as 'true' if the text mentions any weather-related effects on farming and livestock management operations. This includes yield variations in crops and animal products; direct damage to crops, timber resources, or livestock; modifications to agricultural practices or schedules; disruptions to food production or supply chains; impacts on farming equipment and resources; or effects on agricultural inputs including soil conditions, water availability for farming, and essential materials such as seedlings, fertilizers, or animal feed.\\
        
        3. \textbf{Ecological Impact}: Classify as 'true' if the text mentions any effects on natural environments and ecosystems. This includes alterations to local environments and biodiversity; impacts on wildlife populations and behavior patterns; effects on non-agricultural plant life and vegetation; modifications to natural habitats including water bodies, forests, and wetlands; changes in hydrological systems such as river levels and lake conditions; or impacts on urban plant life.\\
        
        4. \textbf{Financial Impact}: Classify as 'true' if the text explicitly mentions economic consequences of weather events. This includes direct monetary losses; business disruptions or closures requiring financial intervention; market price fluctuations or demand changes for specific goods; impacts on tourism and local economic activities; or insurance claims or economic relief measures. Focus only on explicit mentions of financial losses or fluctuations.\\
        
        5. \textbf{Human Health Impact}: Classify as 'true' if the text mentions physical or mental health effects of weather events on populations. This includes direct injuries or fatalities (including cases where zero or more casualties are explicitly mentioned); elevated risks of weather-related or secondary illnesses; mental health consequences such as stress or anxiety; impacts on healthcare service accessibility; or long-term health implications.\\
        
        6. \textbf{Political Impact}: Classify as 'true' if the text mentions governmental and policy responses to weather events. This includes government decision-making and policy modifications in response to events; changes in public opinion or political discourse; effects on electoral processes or outcomes; international relations and aid responses; or debates surrounding disaster preparedness and response capabilities.\\

        Note: \\
        - Return 'false' for any impact category that is either not present in the text or not related to weather events\\
        - Base classifications on explicit mentions in the text\\
        - Focus on direct impacts rather than implications\\
        - Consider immediate and direct effects\\

        Answer only once in the following format:\\
        Infrastructural: true/false\\
        Agricultural: true/false\\
        Ecological: true/false\\
        Financial: true/false\\
        Human Health: true/false\\
        Political: true/false
        }}
\end{tcolorbox}
\caption{Multi-Label Classification instructions template used as the Zero-Shot prompt.}
\label{category_annot}
\end{table*}

\end{document}